\documentclass{article}

\PassOptionsToPackage{authoryear, round}{natbib}
\usepackage[final]{neurips_2025}

\usepackage[utf8]{inputenc}
\usepackage[T1]{fontenc}
\usepackage{hyperref}
\usepackage{url}
\usepackage{booktabs}
\usepackage{amsfonts}
\usepackage{nicefrac}
\usepackage{microtype}
\usepackage[table, dvipsnames]{xcolor}  % Unified xcolor load with all needed options

\usepackage{amsmath}
\usepackage{multirow}
\usepackage{graphicx}
\usepackage{wrapfig}
\usepackage{subcaption}
\usepackage{adjustbox}
\usepackage[absolute,overlay]{textpos}
\usepackage{enumitem}
\usepackage{pifont}
\usepackage[normalem]{ulem}
\usepackage[most,skins,theorems]{tcolorbox}

\newcommand{\cmark}{\textcolor{green}{\ding{51}}}
\newcommand{\xmark}{\textcolor{red}{\ding{55}}}
\tcbset{
  aibox/.style={
    width=\linewidth,
    top=7pt,
    bottom=2pt,
    left=2pt,
    right=2pt,
    colback=blue!6!white,
    colframe=black,
    colbacktitle=black,
    enhanced,
    center,
    attach boxed title to top left={yshift=-0.1in,xshift=0.1in},
    boxed title style={boxrule=0pt,colframe=white,},
  }
}
\newtcolorbox{AIbox}[2][]{aibox,title=#2,label=#1}

\NewDocumentCommand{\lifan}
{ mO{} }{\textcolor{cyan}{\textsuperscript{\textit{lifan}}\textsf{\textbf{\small[#1]}}}}
\NewDocumentCommand{\sagnik}
{ mO{} }{\textcolor{pink}{\textsuperscript{\textit{sagnik}}\textsf{\textbf{\small[#1]}}}}
\NewDocumentCommand{\hao}
{ mO{} }{\textcolor{purple}{\textsuperscript{\textit{hao}}\textsf{\textbf{\small[#1]}}}}
\NewDocumentCommand{\dilek}
{ mO{} }{\textcolor{blue}{\textsuperscript{\textit{dilek}}\textsf{\textbf{\small[#1]}}}}

\newcommand{\norm}[1]{\left\lVert#1\right\rVert}

\title{Reinforcement Learning Finetunes Small Subnetworks in Large Language Models}

\author{Sagnik Mukherjee\hspace{1em}Lifan Yuan\hspace{1em}Dilek Hakkani-T\"ur\hspace{1em}Hao Peng\hspace{1em}\\
University of Illinois Urbana-Champaign \hspace{1em} \\
\texttt{\{sagnikm3,lifan4,dilek,haopeng\}@illinois.edu}\\
}

\begin{document}

\maketitle

\newcommand{\ours}{\textsc{Extreme Gradient Sparsity }}
\begin{abstract}

Reinforcement learning (RL) yields substantial improvements in large language models’ (LLMs) downstream task performance and alignment with human values. 
Surprisingly, such large gains result from updating only a small subnetwork comprising just 5\%-30\% of the parameters, with the rest effectively unchanged.
We refer to this phenomenon as \textit{parameter update sparsity} induced by RL.
It is observed across all 7 widely-used RL algorithms (e.g., PPO, GRPO, DPO) and all 10 LLMs from different families in our experiments.
This sparsity occurs \emph{without} any explicit sparsity-promoting regularizations or architectural constraints. Finetuning the subnetwork alone recovers the test accuracy, and, remarkably, produces a model nearly identical to the one obtained via full finetuning.
The subnetworks from different random seeds, training data, and even RL algorithms show substantially greater overlap than expected by chance. 
Our analysis suggests that this sparsity is \emph{not} due to updating only a subset of layers; instead, nearly all parameter matrices receive similarly sparse updates. Moreover, the updates to almost all parameter matrices are nearly full-rank,
suggesting RL updates a small subset of parameters that nevertheless span almost the full  subspaces that the parameter matrices can represent.
We conjecture that the this update sparsity can be primarily attributed to training on data that is near the policy distribution; 
techniques that encourage the policy to remain close to the pretrained model, such as the KL regularization and gradient clipping, have limited impact. Our code is available at \url{https://github.com/SagnikMukherjee/sparsity_in_rl}.

% on-policy training, which is supported by our experimental results. 

% Our findings provide fresh evidence for recent findings that on-policy RL better preserves the pretrained capabilities compared to off-policy learning with supervised finetuning (SFT), which we find densely update the parameters.

\end{abstract}

\section{Introduction}
\label{sec:intro}

\begin{figure}
    \centering
    \vspace{-1em}
    
    \includegraphics[width=0.8\textwidth]{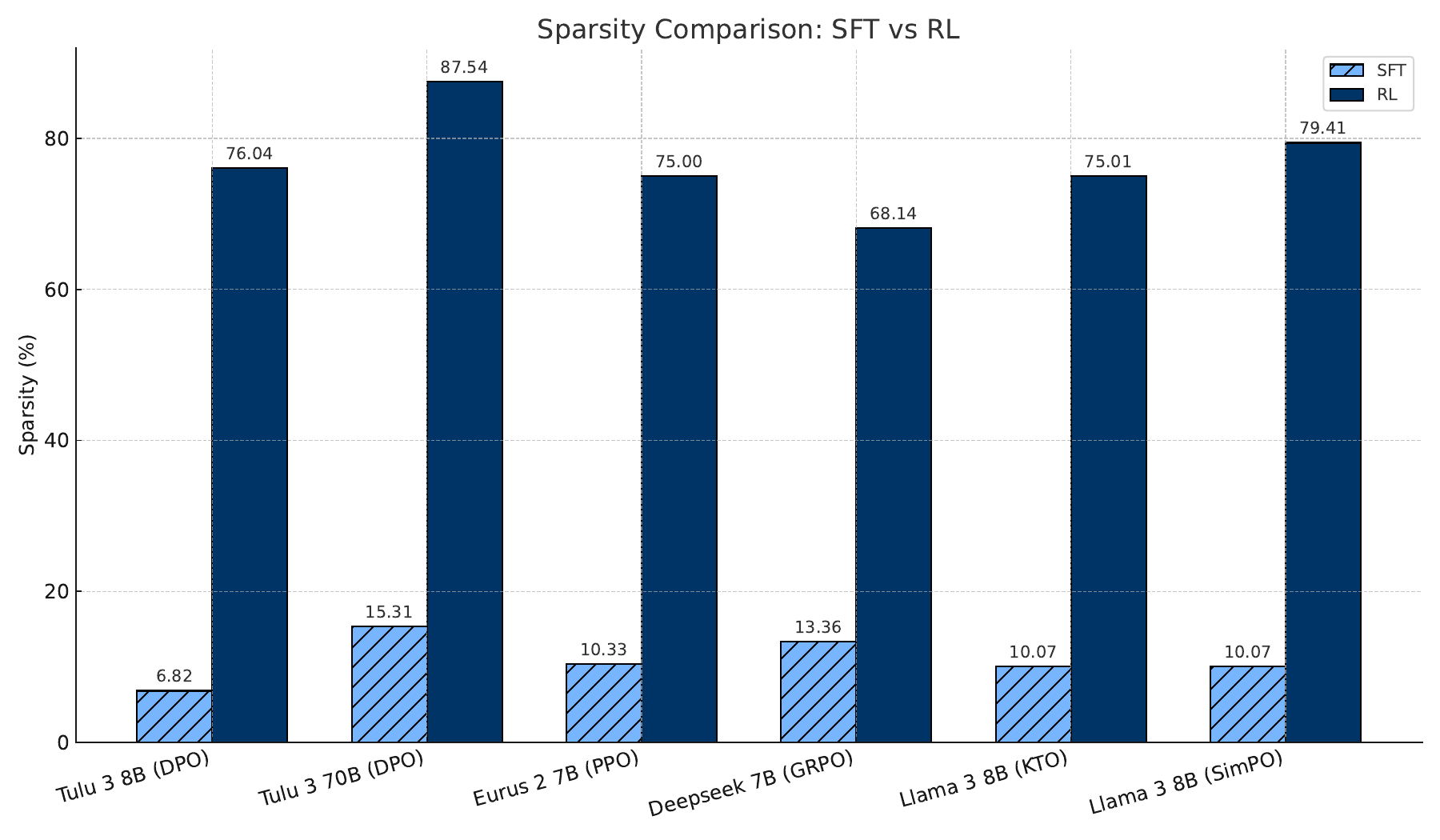}
    \caption{Comparison in accumulated gradients in the SFT stage vs RL stage for popular released checkpoints. SFT stage has accumulated much denser updates, while RL is mostly sparse. }
    \vspace{-.6cm}
    \label{fig:sft-vs-rl}
\end{figure}

% \begin{wrapfigure}{r}{0.50\textwidth}
%     \centering
%     \vspace{-1em}
%     \includegraphics[width=0.50\textwidth]{neurips_2025/figures/sparsity_radar_plot.pdf}
%     \caption*{\small Layerwise Density of accumulated gradient for the Tulu and PRIME 8b models. Here x-axis indicates the index of the particular transformer layer within the model, and y-axis indicates the density. Note that other than the Layer Normalization layer, most layers have similar number of non-zero updates}
%     \vspace{-12em}
% \end{wrapfigure}
% \hao{one thing i'm not happy with: DPO/KTO isn't really considered an "on-policy" RL algo, even if you SFT w/ the positive data.
% we should find alternative wordings instead of on-policy->sparse; off-policy->dense}
% \lifan{but they are both considered as rl, optimizing models at the sequence level. are we really sure that it comes from on-policy data rather than seq-level objective?}
Reinforcement learning (RL) \citep{sutton1998reinforcement, ouyang2022traininglanguagemodelsfollow, ziegler2020finetuninglanguagemodelshuman, ramamurthy2023is, sun-etal-2024-aligning, zhou2025reinforcedmllmsurveyrlbased} is an important post-pretraining stage for adapting large language models (LLMs) to solving complex reasoning problems \citep{lightman2023letsverifystepstep,wang2025otcoptimaltoolcalls,wang2025reinforcementlearningreasoninglarge,cui2025processreinforcementimplicitrewards}, alignment with human values \citep{ouyang2022traininglanguagemodelsfollow,bai2022constitutionalaiharmlessnessai,dai2023saferlhfsafereinforcement}, and adherence to safety protocols \citep{mu2024rulebasedrewardslanguage, huang2024oneshot, zhang2024negative,DBLP:journals/corr/abs-2403-03419}. Since these desired behaviors often differ significantly from those of the pretrained model \citep{ouyang2022traininglanguagemodelsfollow, bai2022constitutionalaiharmlessnessai, OpenAIOS}, it is often assumed that achieving them requires substantial changes to the model’s parameters and therefore full finetuning is widely applied during RL \citep{cui2025processreinforcementimplicitrewards, openr1, liu2025understanding, tinyzero, zeng2025simplerlzooinvestigatingtamingzero}. 
\begin{AIbox}[find:finding]{Finding 1}{\bf RL-induced parameter update sparsity in LLMs: }
RL updates only a small subnetwork of a pretrained large language model, leaving the rest of the parameters effectively unchanged.
\end{AIbox}
%This practice is further reinforced by recent findings that parameter-efficient fine-tuning methods can underperform on challenging reasoning tasks~\citep{biderman2024lora,ivison2023camelschangingclimateenhancing} and potentially lead to undesired behaviors~\citep{shuttleworth2024loravsfinetuningillusion}.
% \newline
% \newline
% \newline
While RL with full finetuning is allowed to update all parameters, does it actually do so?
This paper presents surprising findings and answers this question in the negative.
% \begin{AIbox}{}\textbf{Finding 1:} Reinforcement learning updates only a small subnetwork of a pretrained large language model, leaving the rest of the parameters effectively unchanged.
% \end{AIbox}
Finding \ref{fig:sft-vs-rl} is observed in all 7 widely-used RL algorithms studied, namely PPO \citep{schulman2017proximalpolicyoptimizationalgorithms}, GRPO \citep{shao2024deepseekmathpushinglimitsmathematical}, 
ORPO \citep{hong2024orpomonolithicpreferenceoptimization},
KTO \citep{ethayarajh2024ktomodelalignmentprospect}, DPO \citep{rafailov2023direct}, SimPO \citep{meng2024simposimplepreferenceoptimization} and PRIME \citep{cui2025processreinforcementimplicitrewards}, as well as supervised finetuning with rejection sampling \citep{xiong2025minimalistapproachllmreasoning}, and 10 models in our experiments, with the subnetworks consisting of as little as 5\%  of the model parameters in some cases (\S\ref{sec:empirical_evidence}). 
It emerges \emph{without} any explicit sparsity-promoting regularization, architectural constraint, or use of parameter-efficient training or pruning methods. 
Moreover, we observe a strong consistency among the subnetworks emerged under different random seeds, training data and its order, and even different RL algorithms,
suggesting that the pretrained model contains a partially transferable structure
that is consistent across varied training conditions (\S\ref{sec:generality}).

Interestingly, our experiments with PRIME suggest that approximately 20\% of the parameters are consistently updated and make up the subnetwork. An additional 8\% receive non-zero gradients during training that cancel out, while the remaining $\sim$70\% parameters remain \emph{untouched} throughout the entire training process. This observation motivates us to articulate the following conjecture:
\begin{wrapfigure}{r}{0.40\textwidth}
    \centering
    \includegraphics[width=0.40\textwidth]{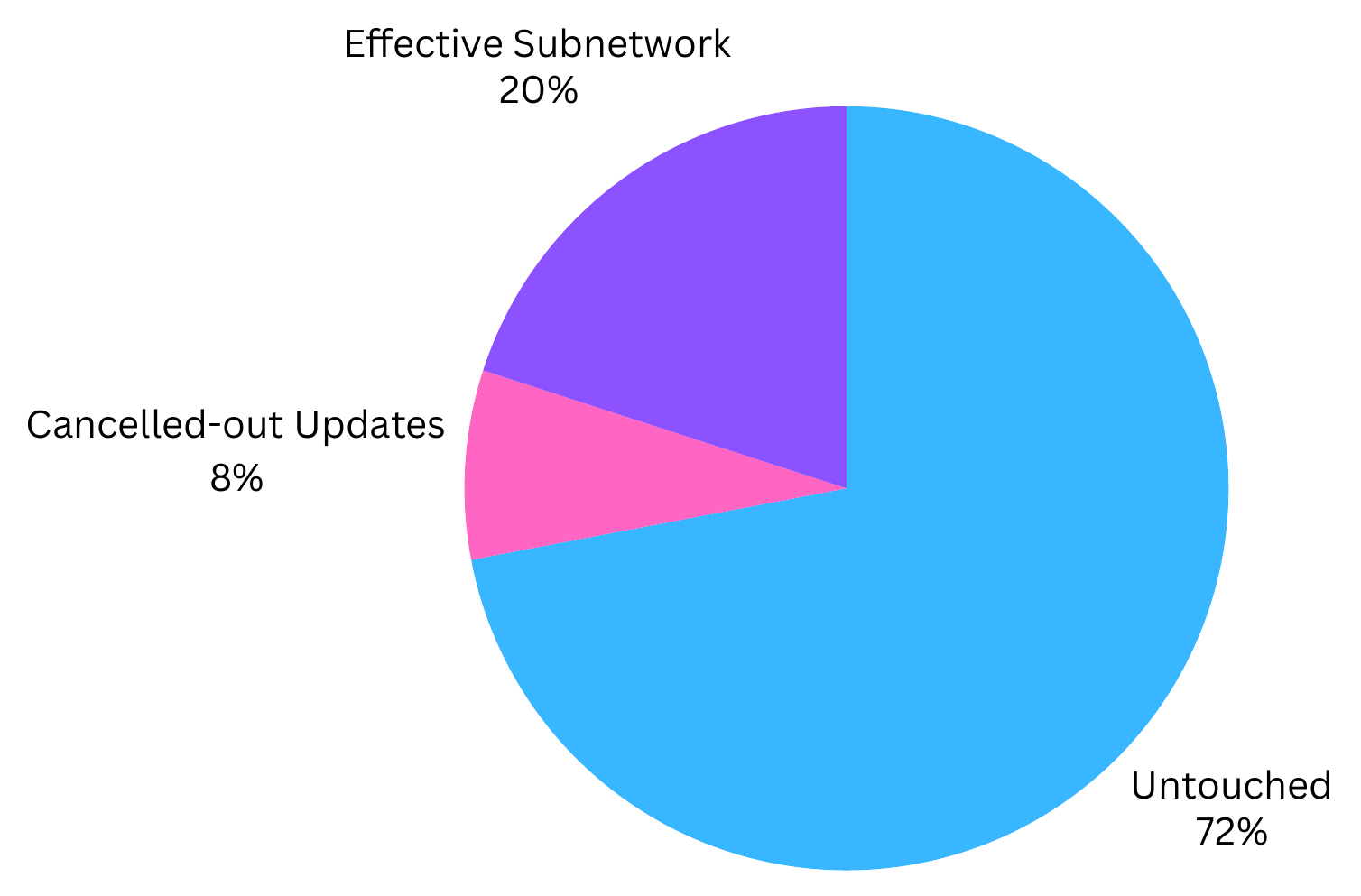}
    % \small
    \caption{In PRIME, 72\% parameters are never updated, 8\% have gradients canceling each other out, and 20\% constitute the subnetwork that is consistently updated (\S\ref{sec:reason})}

    \label{fig:intermediate_sparsity_prime}
    \vspace{-9em}
\end{wrapfigure}
\begin{AIbox}[conj:subnetwork-finetuning]{Conjecture 1}{}\label{conj}
Fine-tuning only the subnetwork identified at the end of RL training, with all other parameters frozen, produces a model that is nearly identical to the original model that has undergone full  finetuning, both in test accuracy and in parameter values.
\end{AIbox}

More formally, let $\theta_{\text{full}}$ denote the parameters after full RL finetuning from the initial model $\theta_{\text{init}}$. 
Define a binary mask $m \in \{0,1\}^{|\theta_{\text{init}}|}$ where $m_i=1$ if $(\theta_{\text{init}}-\theta_{\text{full}})_i\neq 0$ and $0$ otherwise.
We finetune a second model from $\theta_{\text{init}}$ on the same data with the same hyperparameters and number of gradient updates,
but, at each step, mask the gradients as $m \odot \nabla_{\theta} \mathcal{L}(\theta)$  right before the parameter update, so that only the identified subnetwork receives non-zero gradients.
Let $\theta_{\text{sub}}$ denote the resulting parameters of this subnetwork finetuning process. 
Conjecture 1 states that $\theta_{\text{sub}}\approx\theta_{\text{full}}$.
We provide supporting evidence for the conjecture on PRIME and DPO in \S\ref{sec:subnetwork_sufficiency}.

We find that the updates of RL finetuning do  \emph{not} concentrate in specific layers or components of the transformer. Instead, nearly all parameter matrices receive similarly sparse updates (\S\ref{sec:empirical_evidence}). An exception is the layer normalization layers, which receives little to no updates.
Moreover, despite the sparsity in the updates, they are almost always full-rank. 
This suggests that, instead of forcing the updates to reside in a low-rank subspace as in LoRA~\citep{hu2022lora},
RL full finetuning updates a small subset of parameters that nevertheless span almost the full  subspaces that the parameter matrices can represent.

% This suggests that it is impossible for low-rank methods such as LoRA \citep{hu2022lora}
% to fully recover the original full finetuning model in terms of exact model weights.
% This can partially explain why
% forcing the parameter update to be low-rank using LoRA might hurt the performance, as we observe in our experiments. \hao{TODO}

% are ill-suited to capture RL-induced updates which  is supported by empirical results showing that \hao{TODO}.
To better understand the potential reasons for this phenomenon, we conduct a series of experiments in \S\ref{sec:reason}. The results indicate that a primary factor is training on data that is near the policy distribution through, e.g., on-policy RL or performing supervised finetuning (SFT) on the same data before RL \citep{wang2024mathshepherdverifyreinforcellms, cui2025processreinforcementimplicitrewards}.
Intuitively, it requires less change to the policy distribution when the model learns on a sequence sampled from a distribution close to itself. 
In contrast, SFT often involves distribution shifts \citep{zhang2025bestinstructiontuningdatafit} and densely updates the models in our experiments (\S\ref{sec:empirical_evidence}).
Other factors like KL-divergence regularization towards the reference model, gradient clipping (as used in PPO, GRPO, and PRIME), online vs. offline RL, all have limited impact on the sparsity of accumulated updates.

Our findings have important implications for the RL fine-tuning stage of LLMs. They suggest that when RL fine-tuning is performed on data closely aligned with the current policy, as is typical in practice, optimization concentrates primarily on a small, consistently active subnetwork, leaving most other parameters effectively inert.
Conjecture \ref{conj:subnetwork-finetuning} goes beyond the lottery ticket hypothesis (LTH) \citep{frankle2018the}:
not only can the subnetwork, finetuned in isolation, match the performance of the full model in performance as posited by LTH, we show that it also converges to an effectively identical model.
These results offer fresh evidence supporting recent findings that RL better preserves the pretrained capabilities compared to SFT \citep{chu2025sftmemorizesrlgeneralizes,setlur2025scalingtesttimecomputeverification}, potentially by updating substantially fewer parameters.
They also open up new possibilities for more efficient RL training methods that explicitly leverage this update sparsity \citep{chen2022coarseninggranularitystructurallysparse}.

\section{Related work and Background}
% In this section, we provide necessary related work and background for the rest of the paper.
\subsection{Related Work}
\label{sec:related_work}
The Lottery Ticket Hypothesis (LTH; \citealp{frankle2018the}) posited that dense neural networks contain sparse subnetworks capable of matching the performance of the full model when trained in isolation. Subsequent extensions to LLMs identified task-specific subnetworks that mitigate catastrophic forgetting without retraining entire models \citep{panda2024lotteryticketadaptationmitigating, pmlr-v202-panigrahi23a, yadav2023tiesmergingresolvinginterferencemerging}. Related efforts further discovered sparse subnetworks in pretrained language models crucial for encoding specific knowledge \citep{marks2025sparsefeaturecircuitsdiscovering,bayazit2024discoveringknowledgecriticalsubnetworkspretrained,liu2022winwindealsparserobust}. Recent works have also explored exploiting the winning lotteries to improve training efficiency \citep{chen2022coarseninggranularitystructurallysparse}.
While our observation is closely related to LTH, it differs in three core dimensions: (1) LTH identifies winning tickets by pruning, while we study subnetwork that naturally emerge; (2) LTH showed that the final model's performance can be reproduced, we show that, in addition to the performance, the exact same model can almost be recovered;
(3) LTH focuses on models trained from scratch, while we focus on finetuning from pretrained LLMs.

% posits reproducing accuracy of the final model, but we posit convergence of model weights to the same point alongside accuracy, and (3) Lottery Ticket Hypothesis talks about winning tickets that are identified at initialization, i.e. they talk about neural networks trained from scratch, while our investigation is at a much later stage of LLM post-pretraining, the alignment stage.

% Existing literature highlight that reinforcement learning promotes more adaptive reasoning relative to supervised fine-tuning (SFT), which often results in rigid, imitative behaviors and limited generalization capabilities \citep{chen2025sftrlearlyinvestigation, chu2025sftmemorizesrlgeneralizes,li2025reinforcementlearningoutperformssupervised,tan2025reasonrftreinforcementfinetuningvisual}. Recent findings also demonstrate RL's superior scalability at test-time inference \citep{setlur2025scalingtesttimecomputeverification}.

% \textbf{PEFT for alignment/RL: }

Sparse training methods exhibit notable benefits in RL efficiency \citep{graesser2022statesparsetrainingdeep, sidahmed2024parameterefficientreinforcementlearning,sokar2022dynamicsparsetrainingdeep,tan2023rlx2trainingsparsedeep,davelouis2025interplaysparsitytrainingdeep}. Recent studies also employ Low-Rank Adaptation (LoRA) \citep{hu2022lora} in RL and  have achieved competitive performance alongside significantly reduced computational overhead \citep{sidahmed2024parameterefficientreinforcementlearning, wang2025tinatinyreasoningmodels}. 
In contrast to approaches like LoRA that explicitly constrain updates to a small number of parameters, we find that fine-tuning the naturally emerging subnetwork can match or even surpass the performance of full-model finetuning. Moreover, despite its sparsity, the updates are nearly full-rank.
% , suggesting a rich and expressive optimization trajectory.

% It's noteworthy that while these prior works actively seek a sparse subnetwork we observe it intrinsically emgerges from RL, without any sparsity promoting techniques (regularization, pruning or even constraints)

% \hao{besides LTH, we should talk about PEFT for alignment/RL here; it should include lora as well as other approaches}

\subsection{Background}
\label{sec:background}
We briefly introduce key concepts and notations to be used in onward discussion.
% , including on-policy versus off-policy training, and stabilization techniques. We use these concepts throughout the rest of the paper. 

% \hao{revise the wordings here: on-policy is in distribution; off-policy can be in-distribution or off-dist}
% \hao{we need to clarify DPO/KTO (and maybe others) are off-policy algos; 
% when one SFT on the positive data (common practice), the RL phase becomes more in-distribution.
% when it is not SFT-ed first, it is off-distribution
% on-policy RL algos are in-distribution.
% }
\noindent\textbf{The sparsity of parameter updates.} Let $\theta^0, \theta^1 \in \mathbb{R}^n$ denote the model parameters before and after finetuning, respectively. 
We define the \textbf{update sparsity} as $\text{sparsity}(\theta^0, \theta^1) := 1 - \norm{\theta^1 - \theta^0}_0/n$, 
where $\norm{\cdot}_0$ counts the number of non-zero elements. 
It is important to clarify that even the update $\theta_1-\theta_0$ is sparse, it does not imply that the finetuned model $\theta_1$ is sparse.
Since no sparsity is assumed for $\theta_0$, a sparse update can still result in a dense $\theta_1$ if $\theta_0$ is dense.

Unless otherwise specified, we follow standard practice and consider two \texttt{bfloat16} values as equal when their absolute difference does not exceed $10^{-5}$, to account for numerical precision limits.\footnote{\href{https://docs.pytorch.org/docs/stable/generated/torch.autograd.gradcheck.gradcheck.html}{E.g., PyTorch uses $10^{-5}$ as the default tolerance for gradient checking.}} 
All models in our experiments are in the \texttt{bfloat16} data type.
Sparsity with different tolerance values can be found in Table \ref{tab:full_sparsity_table} in the Appendices.

% \hao{i trimmed this. check}
% parameters updated during fine-tuning.

\noindent\textbf{Learning from in-distribution data.} 
We use “in-distribution” to refer to training on data drawn from a distribution that closely matches the current policy. 
An example is on-policy RL with, e.g., PPO, GRPO, and PRIME, which sample data online from the evolving policy during training. 
Another way to achieve in-distribution RL is to perform SFT on the same data used for subsequent RL, so that the policy adapts to the data distribution before RL.
This is a common practice in off-policy methods like DPO and KTO. 
On-policy methods inherently train on in-distribution data, and off-policy methods can also do so when the training data closely matches the policy distribution.
As we will show later in \S\ref{sec:reason}, training on in-distribution data is a primary reason for the update sparsity in RL.

% the base policy; sequences with high likelihood under the current policy $\pi_\theta$ is considered In-distribution. In contrast, Out-of-distribution algorithms operate on fixed datasets, often collected from different policies, which are not necessarily aligned with the current model's distribution. 
% SFT, by nature, trains on Out-of-distribution data. On policy RL algorithms such as PPO and GRPO as well as off-policy methods such as DPO/KTO train on In-Distribution samples, while vanilla SFT often uses Out-of-distribution data. Note that, on-policy methods are all strictly in-distribution, since they sample from the policy itself. However off-policy ones can be completely out of distribution (such as vanilla SFT) or near-on-policy such as iterative DPO\citep{pang2024iterativereasoningpreferenceoptimization}.

\noindent\textbf{KL-divergence regularization and gradient clipping in RL.}
Two widely adopted techniques to keep the policy from deviating too far from the reference model are KL-divergence regularization and gradient clipping.
KL regularization \citep{schulman2017trustregionpolicyoptimization}, formally computed as $D_{\text{KL}}(\pi_{\theta} \| \pi_{\text{ref}}) = \mathbb{E}_{\pi_\theta}\left[\log \frac{\pi_{\theta}(y|x)}{\pi_{\text{ref}}(y|x)}\right]$, constrains policy shifts. Gradient clipping further stabilizes training by bounding the update norm.
Both are widely used in algorithms such as PPO, GRPO, and PRIME.
In \S \ref{sec:reason}, we show that, counterintuitively, both have limited impact on the  update sparsity.

\section{RL Induces Sparse but Full-rank Updates; SFT Induces Dense Ones}
\label{sec:empirical_evidence}
% \hao{needs to be better momtivated:}
This section aims to answer the following research question

\begin{itemize}[noitemsep,topsep=0pt,parsep=0pt,partopsep=0pt]
\item[\textbf{RQ1:}] \textit{To what extent does RL induce sparse parameter updates and where in the model do these updates occur? How does SFT compare?} 
\end{itemize}

% The performance gap between on‑policy and off‑policy methods is commonly ascribed to distributional shifts \citep{xu2024is, tang2024understandingperformancegaponline}, and a frequent remedy is to align training data with the policy distribution \citep{guo2024directlanguagemodelalignment,zhang2024textbfplumimprovingcodelms,xiong2024iterative,zhuang2023behavior}. When data already consist of high‑likelihood samples under the base model, however, resulting policy updates may be attenuated. We investigate this effect through a mechanistic analysis of the accumulated gradients by asking the following question: \textbf{RQ1: }\textit{Does RL induce sparse parameter updates in LLMs? If so, to what extent and where in the model are these updates? How does SFT fare?} 

% \hao{i suggest separating the discussion of RL and SFT:
% para 1: RL is sparse, para 2: RL is dense, so that we don't need to get back and forth on the checkpoints we are using
% }\lifan{i suggested merging because our discussion on sft is very shallow, merging them allows us to only talk about rl which looks more in-depth}
% \hao{i don't see why it is necessary to use the $\theta_0$ and $\theta_1$ notations in this section}

\noindent\textbf{Setup.}  
To answer this question, we analyze publicly released model checkpoints on Hugging Face released by the authors. 
With the exception of models where RL is applied directly to the pretrained base model (e.g., \texttt{DeepSeek-R1-Zero}), most models follow a conventional three-stage pipeline: pretraining, supervised fine-tuning (SFT), and RL. 
We analyze both the RL and SFT stages by measuring the update sparsity between model checkpoints before and after RL or SFT fine-tuning.
Our experiments cover \texttt{Tulu 8B/70}B~\citep{lambert2025tulu3pushingfrontiers}, \texttt{Eurus 7B}~\citep{yuan2024advancing, cui2025processreinforcementimplicitrewards}, \texttt{DeepSeek Math 7B}~\citep{shao2024deepseekmathpushinglimitsmathematical}, and KTO/SimPO models~\citep{meng2024simposimplepreferenceoptimization}.

\begin{table}[t]
\centering
% \small
\caption{Parameter update sparsity  across different RL algorithms. We report sparsity for a suite of open models from Hugging Face. For all models, at least 68.5\%---and often much more---of the parameters remain unchanged after RL.}
\adjustbox{max width=\textwidth}{
\begin{tabular}{@{} lllcccc @{} }
\toprule
\textbf{Algo.} & \textbf{Init Model} & \textbf{RL Model} & \textbf{Update Sparsity} & \textbf{On-Policy} & \textbf{KL} & \textbf{Online} \\
\midrule
\multirow{2}{*}{DPO} 
 & \href{https://huggingface.co/allenai/Llama-3.1-Tulu-3-8B-SFT}{\texttt{Llama-3.1-Tulu-3-8B-SFT}} & \href{https://huggingface.co/allenai/Llama-3.1-Tulu-3-8B-DPO}{\texttt{Llama-3.1-Tulu-3-8B-DPO}} & 81.4 & \xmark & \cmark & \xmark \\
 & \href{https://huggingface.co/allenai/Llama-3.1-Tulu-3-70B-SFT}{\texttt{Llama-3.1-Tulu-3-70B-SFT}} & \href{https://huggingface.co/allenai/Llama-3.1-Tulu-3-70B-DPO}{\texttt{Llama-3.1-Tulu-3-70B-DPO}} & 95.2 & \xmark & \cmark & \xmark \\
\midrule
\multirow{2}{*}{GRPO} 
 & \href{https://huggingface.co/deepseek-ai/deepseek-math-7b-instruct}{\texttt{deepseek-math-7b-instruct}} & \href{https://huggingface.co/deepseek-ai/deepseek-math-7b-rl}{\texttt{deepseek-math-7b-rl}} & 68.5 & \cmark & \cmark & \cmark \\
 & \href{https://huggingface.co/deepseek-ai/DeepSeek-V3-Base}{\texttt{DeepSeek v3 base}} & \href{https://huggingface.co/deepseek-ai/DeepSeek-R1-Zero}{\texttt{DeepSeek-R1-Zero}} & 86.0 & \cmark & \cmark & \cmark \\
\midrule
ORPO & \href{https://huggingface.co/mistralai/Mistral-7B-v0.1}{\texttt{mistral-7B-v0.1}} & \href{https://huggingface.co/kaist-ai/mistral-orpo-beta}{\texttt{mistral-orpo-beta}} & 76.9 & \xmark & \xmark & \xmark \\
\midrule
\multirow{2}{*}{KTO} 
 & \href{https://huggingface.co/openbmb/Eurus-7b-sft}{\texttt{Eurus-7b-sft}} & \href{https://huggingface.co/openbmb/Eurus-7b-kto}{\texttt{Eurus-7b-kto}} & 96.0 & \xmark & \cmark & \xmark \\
 & \href{https://huggingface.co/princeton-nlp/Llama-3-Base-8B-SFT}{\texttt{Llama-3-Base-8B-SFT} }& \href{https://huggingface.co/princeton-nlp/Llama-3-Base-8B-SFT-KTO}{\texttt{Llama-3-Base-8B-SFT-KTO}} & 81.2 & \xmark & \cmark & \xmark \\
\midrule
PPO & \href{https://huggingface.co/peiyi9979/mistral-7b-sft}{\texttt{mistral-7b-sft}} & \href{https://huggingface.co/peiyi9979/math-shepherd-mistral-7b-rl}{\texttt{math-shepherd-mistral-7b-rl}} & 80.8 & \cmark & \cmark & \cmark \\
\midrule
SimPO & \href{https://huggingface.co/meta-llama/Meta-Llama-3-8B-Instruct}{\texttt{Meta-Llama-3-8B-Instruct}} & \href{https://huggingface.co/princeton-nlp/Llama-3-Instruct-8B-SimPO}{\texttt{Llama-3-Instruct-8B-SimPO}} & 86.5 & \xmark & \xmark & \xmark \\
\midrule
PRIME & \href{https://huggingface.co/PRIME-RL/Eurus-2-7B-SFT}{\texttt{Eurus-2-7b-sft}} & \href{https://huggingface.co/PRIME-RL/Eurus-2-7B-PRIME}{\texttt{Eurus-2-7B-PRIME}} & 77.0 & \cmark & \xmark & \cmark \\
\bottomrule
\end{tabular}}
\label{tab:gross_sparsity}
    % \vspace{-.5cm}
\end{table}

% \begin{table}[h]
% \centering
% \small
% \caption{Sparsity in net accumulated gradient in the RL training stage across a suite of open-weight models from huggingface. Note that most models have at least 70\% weights that did not undergo any change.}
% \adjustbox{max width=\textwidth}{
% \begin{tabular}{@{} lllllll @{} }
% \toprule
% \textbf{Algo.} & \textbf{Init Model} & \textbf{RL Model} & \textbf{ Sparsity in Updates(\%)} & \textbf{Policy Type} & \textbf{KL} & \textbf{Base Policy}\\
% \midrule
% \multirow{ 2}{*}{DPO}
%  & \href{https://huggingface.co/allenai/Llama-3.1-Tulu-3-8B-SFT}{Llama-3.1-Tulu-3-8B-SFT} & \href{https://huggingface.co/allenai/Llama-3.1-Tulu-3-8B-DPO}{Llama-3.1-Tulu-3-8B-DPO} & 81.4 \\
%     & \href{https://huggingface.co/allenai/Llama-3.1-Tulu-3-70B-SFT}{Llama-3.1-Tulu-3-70B-SFT} & \href{https://huggingface.co/allenai/Llama-3.1-Tulu-3-70B-DPO}{Llama-3.1-Tulu-3-70B-DPO} & 95.2 \\
% \midrule
% \multirow{ 2}{*}{GRPO} & \href{https://huggingface.co/deepseek-ai/deepseek-math-7b-instruct}{deepseek-math-7b-instruct} & \href{https://huggingface.co/deepseek-ai/deepseek-math-7b-rl}{deepseek-math-7b-rl} & 68.5 \\
%      & \href{https://huggingface.co/deepseek-ai/DeepSeek-V3-Base}{Deepseek v3 base} & \href{https://huggingface.co/deepseek-ai/DeepSeek-R1-Zero}{DeepSeek-R1-Zero} & 86.0 \\
% \midrule
% ORPO & \href{https://huggingface.co/mistralai/Mistral-7B-v0.1}{Mistral-7B-v0.1} & \href{https://huggingface.co/kaist-ai/mistral-orpo-beta}{mistral-orpo-beta} & 76.9 \\
% \midrule
% \multirow{ 2}{*}{KTO} & \href{https://huggingface.co/PRIME-RL/Eurus-2-7B-SFT}{Eurus-2-7b-sft} & \href{https://huggingface.co/openbmb/Eurus-7b-kto}{Eurus-7b-kto} & 96.0 \\
%     & \href{https://huggingface.co/princeton-nlp/Llama-3-Base-8B-SFT}{Llama-3-Base-8B-SFT} & \href{https://huggingface.co/princeton-nlp/Llama-3-Base-8B-SFT-KTO}{Llama-3-Base-8B-SFT-KTO} & 81.2 \\
% \midrule
% PPO & \href{https://huggingface.co/peiyi9979/mistral-7b-sft}{mistral-7b-sft} & \href{https://huggingface.co/peiyi9979/math-shepherd-mistral-7b-rl}{math-shepherd-mistral-7b-rl} & 80.8 \\
% \midrule
% SimPO & \href{https://huggingface.co/meta-llama/Meta-Llama-3-8B-Instruct}{Meta-Llama-3-8B-Instruct} & \href{https://huggingface.co/princeton-nlp/Llama-3-Instruct-8B-SimPO}{Llama-3-Instruct-8B-SimPO} & 86.5 \\
% \midrule
% PRIME & \href{https://huggingface.co/PRIME-RL/Eurus-2-7B-SFT}{Eurus-2-7b-sft} & \href{https://huggingface.co/PRIME-RL/Eurus-2-7B-PRIME}{Eurus-2-7B-PRIME} & 77.0 \\
% \bottomrule
% \end{tabular}}
% \label{tab:gross_sparsity}
% \end{table}

\noindent\textbf{Results.}
As shown in Table \ref{tab:gross_sparsity}, for all RL-finetuned models, 68.5\%–96.0\% of parameters remain unchanged after RL. This trend holds across different RL algorithms and model families. 
Particularly, \texttt{Deepseek-R1-Zero} presents a update sparsity of 86.0\%, regardless of directly training from the pretrained base model, namely RL-Zero \citep{deepseekai2025deepseekr1incentivizingreasoningcapability}, and large-scale training for over 8K steps.
Although exact training configurations are not always available, we observe that within the same model family, larger models tend to show higher sparsity.
% \hao{i added below, please check:}
Importantly, all of these models are trained using full finetuning \emph{without} any sparsity-promoting regularization techniques or constraints.
This suggests that the update sparsity emerges naturally.

In contrast, Figure~\ref{fig:sft-vs-rl} shows that SFT induces dense updates (only 6\%-15\% sparsity).
These results offer fresh evidence supporting recent findings that RL better preserves the pretrained capabilities than SFT \citep{chu2025sftmemorizesrlgeneralizes,setlur2025scalingtesttimecomputeverification}, possibly by updating substantially fewer parameters.

% This suggests that SFT produces dense parameter updates, whereas RL tends to operate in a highly sparse regime.

\begin{AIbox}[]{Takeaway 1}{}
RL leads to consistently sparse parameter updates (often $>$70\% sparsity) while SFT produces dense updates. This sparsity emerges without regularizations or architectural constraints.
\end{AIbox}
% \lifan{the logics in this section is a bit messy. we've been frequently switching between results and setups. and we need some boled text to guide readers how to quickly parse info from the three paragraphs after "how does sft fare".}

% \dilek{clarify that a tolerance value was used}
% We begin by providing empirical evidence for the gradient sparsity for RL training. For doing so, we analyzed publicly released model checkpoints from Hugging Face, focusing on several prominent RL algorithms, including DPO, KTO, PPO, GRPO, PRIME, and others. 
% We are following our previous definition in Eq. \ref{eq:density_sparsity}. Here $\theta^1$ refers to an RL trained model, and $\theta^0$ points to the model where training was initiated from. 

% \hao{the following two paragraphs can be merged into one:}
% \noindent\textbf{RL finetuning sparsely updates the parameters.}

% \begin{figure}[t]
%   \centering
%   \includegraphics[width=0.75\textwidth]{neurips_2025/figures/sft_vs_rl.pdf}
%   \caption{Sparsity comparison between SFT and RL stages. Each bar indicates the proportion of parameters not updated during each phase. \lifan{eurus 2 prime? not ppo?}}
%   \label{fig:sft-vs-rl}
% \end{figure}

\begin{figure}[t!]
    \centering
    
    \includegraphics[width=\textwidth]{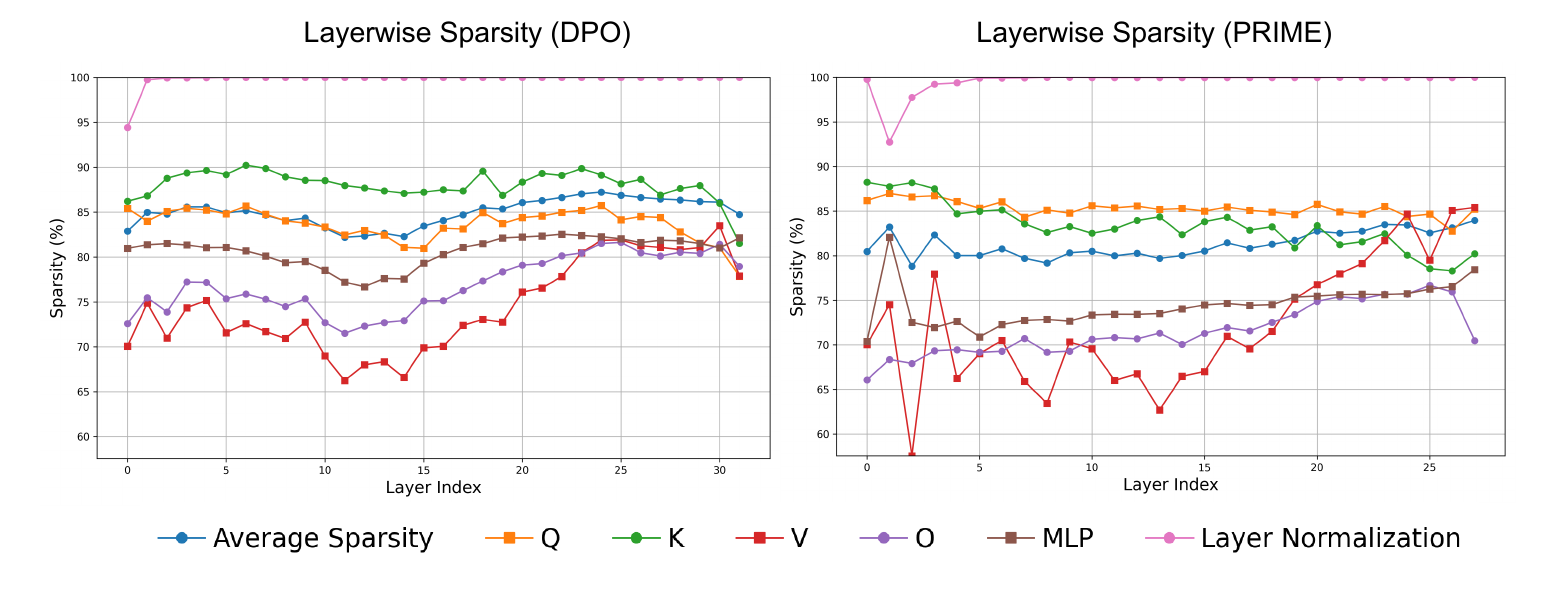}
    \vspace{-.75cm}
    \caption{Layerwise and per-parameter-matrix update sparsity for DPO (left) and PRIME (right). All layers are similarly sparsely updated, with the only exception of the layer normalization layers, which receive little to no updates.}
    \label{fig:layerwise_sparsity}
    \vspace{-.5cm}
\end{figure}
\label{sec:layerwise}
% \hao{did we point to the figures}
% \hao{be clear this one is about rl}
% \hao{i rewrote this one. check}
\begin{wraptable}{r}{0.45\textwidth}
    \centering
    \small
    \vspace{-1em}
    \caption{Mean ranks of update matrices, as a percentage of maximum possible rank across models after RL finetuning.}
    \setlength{\tabcolsep}{4pt}
    \renewcommand{\arraystretch}{1.2}
    \begin{tabular}{@{} lc@{}}
        \toprule
        \textbf{Model and Algo.} & \textbf{Update Rank (\%)} \\
        \midrule
        \href{https://huggingface.co/allenai/Llama-3.1-Tulu-3-8B-DPO}{Tulu 8B (DPO)} & 99.8 \\
        \href{https://huggingface.co/PRIME-RL/Eurus-2-7B-PRIME}{Eurus 7B (PRIME)} & 99.5 \\
        \href{https://huggingface.co/princeton-nlp/Llama-3-Base-8B-SFT-KTO}{Llama-3 8B (KTO)} & 99.2 \\
        \href{https://huggingface.co/deepseek-ai/deepseek-math-7b-rl}{DeepSeek Math 7B (GRPO)} & 99.4 \\
        \bottomrule
    \end{tabular}
    
    \vspace{-1em}
    \label{tab:updates_rank}
\end{wraptable}
\noindent\textbf{Almost all transformer layers receive similarly sparse updates.}
We next examine how parameter updates in RL are distributed across the model layers and individual parameter matrices (e.g., Q, K, V projections), based on DPO and PRIME models.
If updates \emph{were} concentrated in a subset of layers or modules, one could exploit that structure for enhancing the efficiency \citep{pan2024lisa}. 
Figure~\ref{fig:layerwise_sparsity} shows layerwise and per-matrix sparsity across the models.
The "Average Sparsity" is over each transformer layer, while others correspond to specific parameter matrices.
We observe that parameter updates are distributed across different matrices rather than localized to specific ones.
Except for consistently high sparsity in layer normalization layers, most layers exhibit similar sparsity levels.
% In addition, the sparsity of the updates to the embedding and output projection layers are $92.0\%$ and $94.2\%$ in \textsc{Prime} and $82.8\%$ and $77.5\%$ in \textsc{DPO} \lifan{let's attribute it back to the overall sparsity, like how many of the 23\% updates in prime are due to embeddings}. 
% With the exception of consistently high sparsity in LayerNorms, most subcomponents exhibit a fairly uniform sparsity profile, indicating that non-zero updates are distributed throughout the model.
Our results show that sparsity is relatively even across the model. 
This suggests that recovering the behavior of the fully finetuned model requires updating all layers, albeit with only a subset of parameters in each.

\noindent\textbf{Updates are sparse but full-rank.}
Given the sparsity of RL-induced updates, a natural question is whether these updates are also low-rank. 
This distinction between low-rank and sparse updates is important: 
the former would imply that finetuning operates within a subspace, 
while the latter implies that a small subset of parameters (that can span the full parameter space) are selected to finetune.
% while full-rank sparse updates suggest more distributed adaptation across the parameter space. 
Notably, while the updates are sparse, 
a closer inspection reveals that they are nearly full rank (Tab \ref{tab:updates_rank}).
% upon a closer investigation, we observe that the updates are still full rank (see Tab \ref{tab:updates_rank})\hao{in order to say that they are (almost)full rank, we need to show that the min is almost full-rank}, 
% This implies that while sparse, gradient updates are quite informative\hao{any evidence show correlations between rank and informativeness?} in RL. 
To compute rank, we calculate the average rank of individual update matrices across all layers.  
We further examine the rank of the update for each layer and parameter matrix, 
and find that most are full-rank throughout the model. 
These findings suggest that RL updates are localized to a subset of the parameters that almost span the full subspaces that the parameter matrices can represent,
instead of residing in a low-rank subspace.
% and, in this sense, are structurally rich.
% This also implies that under fixed parameter budget, finetuning the subnetwork has higher likelihood of reproducing the full model than training with LoRA.

\begin{AIbox}{Takeaway 2}{}
All layers and parameter matrices receive similarly sparse but full-rank updates. While layer normalization parameters are almost never updated.
\end{AIbox}

\section{Finetuning the Subnetwork Alone Can Reproduce the Full-finetuned Model}
\label{sec:subnetwork_sufficiency}
% \lifan{again, the transition is weird. it seems that you are following your internal thought flow but readers still don't know why they should read the following rather than closing the paper. they don't know why it's a natural extension thought and why it's interesting at all.}
% \hao{agreed. this needs to be better motivated.
% the transition question, which is also the theme of this section, should be: "if rl finetunes only a subnetwork, can we recover fullfinetuning by training that subnetwork alone?"
% }
% \textbf{Research question and Motivation: }
Since RL primarily fine-tunes a small subnetwork, we investigate two research questions inspired by but extending beyond the Lottery Ticket Hypothesis (LTH):
\begin{itemize}[noitemsep,topsep=0pt,parsep=0pt,partopsep=0pt]
\item[\textbf{RQ2:}] \textit{Can finetuning the subnetwork in isolation recover the performance of the full-finetuned model?}
\item[\textbf{RQ3:}] \textit{Can subnetwork-only finetuning also recover the exact parameter values produced by full RL finetuning?}
This section answers both in the positive.
\end{itemize}

% Now that we have established that RL fine-tuning primarily modifies a small subnetwork within the LLM, a key question arises: \textit{If RL primarily fine-tunes a small subnetwork, is it possible to reproduce the full fine-tuned model (exact model weights) by training only that subnetwork in isolation?}. A positive finding would suggest that the remaining parameters play little role in the optimization process, opening up possibilities for more efficient RL training methods that explicitly leverage this sparsity. 

% surprisingly we find that the subnetwork when trained in isolation, can reproduce the full finetuned model to a large extent. Further we observe a model improvent in end task accuracy, as already proposed in LTH. 

\noindent\textbf{Setup. }% We proceed in three steps, briefly summarized here for clarity. 
% We first train the model with full RL finetuning, obtaining the final parameters $\theta_{\text{full}}$ from the initialization $\theta_{\text{base}}$. Next, we identify the updated subnetwork by computing a binary mask $m \in \{0,1\}^{|\theta_{\text{base}}|}$, where $m_i = 1$ if $(\theta_{\text{final}} - \theta_{\text{base}})_i \ne 0$, and $0$ otherwise. 
% Finally, we train a second model starting from $\theta_{\text{base}}$, masking the gradients: $M \odot \nabla_{\theta} \mathcal{L}(\theta)$ before updating the parameters, so that only the identified subnetwork receives updates.
% Lastly, we compare the second model with $\theta_{\text{final}}$ 
% task performance and the values of the parameters.\hao{we had the same thing in the intro.
% it's fine to either drop this one or keep both, but we need to make sure the notations are consistent
% }
% in terms of model weights and final task accuracy to validate our hypothesis.
We follow the procedure described in \S\ref{sec:intro} to obtain two models: one with full finetuning  $\theta_{\text{full}}$, and another finetuned on the same data and hyperparameters but updating only the subnetwork $\theta_{\text{sub}}$.
We experiment on two very different algorithms DPO, an off-policy algorithm using implicit outcome rewards, and PRIME, an on-policy one with process reward models, to ensure that our conclusion can generalize. 
We implement DPO with Open-Instruct and PRIME with verl. The exact hyperparameter choices for both can be found in Appendix \ref{sec:hyperparam_gradient_masking}. For evaluation, we choose a subset of tasks reported in the original papers for both. For DPO we choose the LSAT \citep{10.1109/TASLP.2022.3164218}, LogiQA \citep{10.5555/3491440.3491941} splits from AGIEval \citep{zhong-etal-2024-agieval}, Math split of MMLU Pro \citep{wang2024mmlupro}. For PRIME, we report results on the MATH500 \citep{hendrycks2021measuring} benchmark across difficulty levels. For evaluation in DPO we use olmes.\footnote{Open-Instruct: \url{https://github.com/allenai/open-instruct}; verl: \url{https://github.com/volcengine/verl/}; olmes: \url{https://github.com/allenai/olmes/}}
\begin{wrapfigure}{r}{0.60\textwidth}
    \centering
    \vspace{-1.5em}
    % \vspace{--2}
    \includegraphics[width=0.60\textwidth]{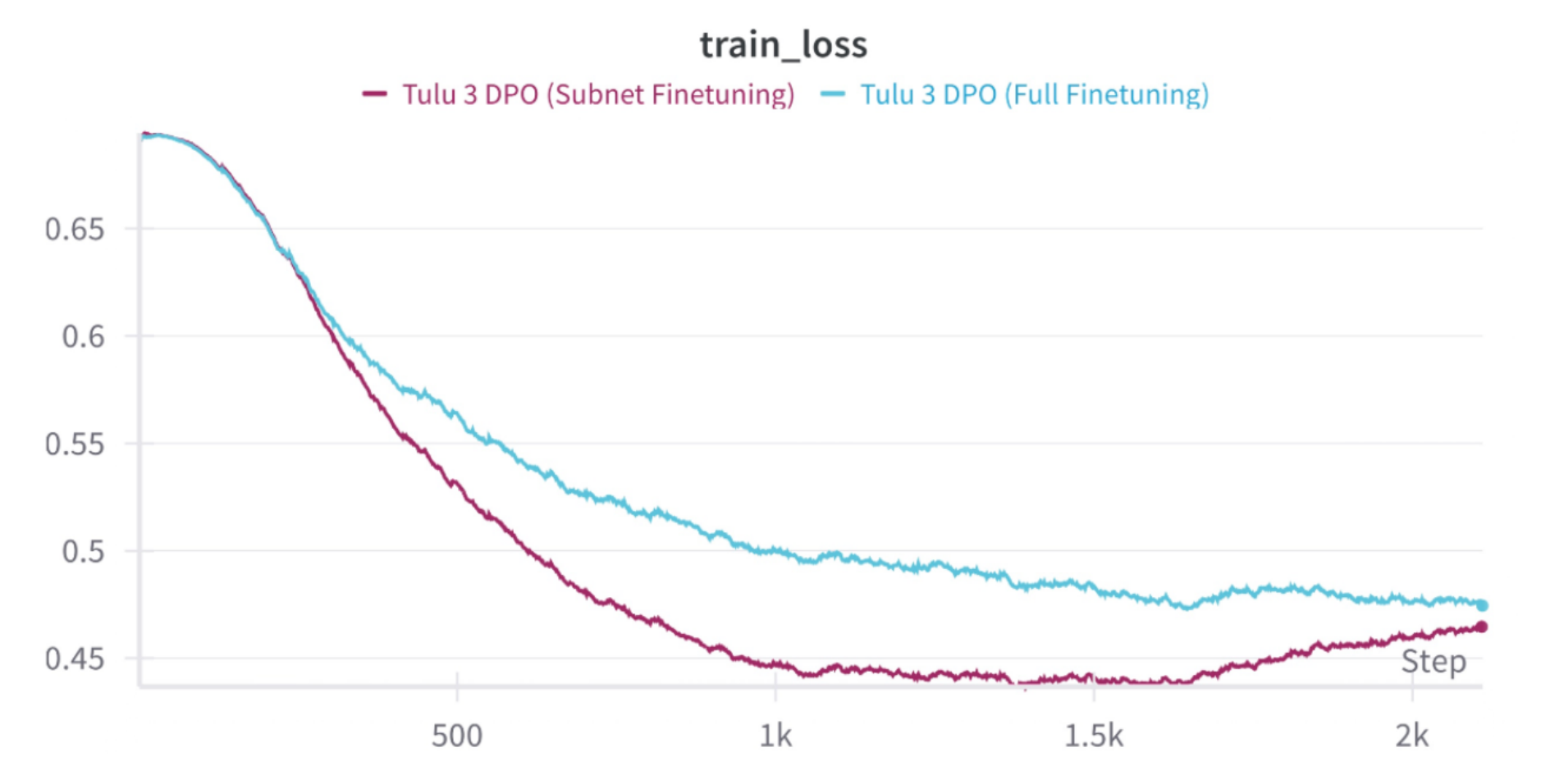}
    \caption{Training loss for training DPO with subnetwork finetuning and full finetuning. Training the subnetwork in isolation consistently causes train loss to be lower.}
    \label{fig:loss_comparison}
    \vspace{-1.3em}
\end{wrapfigure}
\noindent\textbf{Results.} 
% \begin{wrapfigure}{r}{0.50\textwidth}
%     \centering
%     \includegraphics[width=0.50\textwidth]{neurips_2025/figures/loss.pdf}
%     \caption{Update sparsity of intermediate checkpoints of a training run of PRIME. We observe that with more training the sparsity slowly decays}
%     \label{fig:intermediate_sparsity_prime}
% \end{wrapfigure}
In DPO, 94.0\% weights are same between $\theta_{\text{full}}$ and $\theta_{\text{sub}}$;
it is 90.5\% for PRIME.
Notably, for both DPO and PRIME, $\theta_{\text{full}}$ and $\theta_{\text{sub}}$ are 100\% identical when using a tolerance of $10^{-4}$ instead of the default $10^{-5}$, indicating that the two models converge to nearly identical parameter values.
As shown in Tables~\ref{tab:gradient_masking_DPO} and~\ref{tab:gradient_masking_PRIME}, $\theta_{\text{sub}}$ matches or outperforms $\theta_{\text{full}}$ on all tasks across both algorithms. These results suggest that the
parameters outside the subnetwork play little role in the optimization process, and freezing them has a negligible or even beneficial impact on the model's performance. We further observe that the training loss is consistently lower in the subnetwork finetuning setting than full finetuning (Fig. \ref{fig:loss_comparison}).
They also provide supporting evidence for our Conjecture~\ref{conj} in \S\ref{sec:intro}.
This finding opens up new possibilities for more efficient RL training methods that explicitly leverage this update sparsity \citep{chen2022coarseninggranularitystructurallysparse}.

% We also analyze the final task accuracies, as reported in Table \ref{tab:gradient_masking_DPO} and \ref{tab:gradient_masking_PRIME}. We observe a consistent modest improvement in end task accuracy. We hypothesize that this has profound implications in the notorious challenge of compute requirement in RL by reducing FLOPs \citep{chen2022coarseninggranularitystructurallysparse}. 

\begin{table}[h]
  % \small
  \centering
  \caption{Test set performance of $\theta_{\text{full}}$ and $\theta_{\text{sub}}$ trained with DPO and PRIME. Training only the subnetwork ($\theta_{\text{sub}}$) can achieve better performance than full finetuning ($\theta_{\text{full}}$). Lvl. indicates the difficulty levels of MATH500.}
  \begin{subtable}[t][6cm][t]{0.48\textwidth}
    \centering
    \caption{DPO}
    \setlength{\tabcolsep}{6pt}
    \renewcommand{\arraystretch}{1.1}
    \begin{tabular}{lccc}
      \toprule
      \textbf{Task} & $\theta_{\text{full}}$ & $\theta_{\text{sub}}$ & $\Delta$ \\
      \midrule
      AGIEval LSAT-AR      & 21.3 & 24.8 & +3.5 \\
      AGIEval LSAT-LR      & 53.1 & 54.7 & +1.6 \\
      AGIEval LogiQA-en     & 43.5 & 45.5 & +2.0 \\
      GPQA                    & 32.8 & 32.8 & +0.0 \\
      MMLU Pro Math           & 50.8 & 51.6 & +0.8 \\
      \midrule
      \rowcolor{gray!20}
      \textbf{Avg}            & 40.3 & 41.9 & +1.6 \\
      \bottomrule
    \end{tabular}
    \label{tab:gradient_masking_DPO}
  \end{subtable}%
  \hfill
  \begin{subtable}[t][6cm][t]{0.48\textwidth}
    \centering
    \caption{PRIME}
    \setlength{\tabcolsep}{6pt}
    \renewcommand{\arraystretch}{1.1}
    \begin{tabular}{lccc}
      \toprule
      \textbf{Lvl.} & $\theta_{\text{full}}$ & $\theta_{\text{sub}}$ & $\Delta$ \\
      \midrule
      1 & 93.0  & 93.0  & +0.0 \\
      2 & 85.6  & 85.6  & +0.0 \\
      3 & 82.9  & 83.8  & +0.9 \\
      4 & 71.1  & 74.2  & +3.1 \\
      5 & 40.3  & 45.5  & +5.2 \\
      \midrule
      \rowcolor{gray!20}
      \textbf{Overall} & 69.8 & 72.2 & +2.4 \\
      \bottomrule
    \end{tabular}
    \label{tab:gradient_masking_PRIME}
  \end{subtable}
  \vspace{-4em}
  
  \vspace{-2.5em}
  \label{tab:side_by_side_gradient_masking}
\end{table}

\section{Consistency of Subnetworks Across Seeds, Data, and Algorithms}
\label{sec:generality}
% \hao{again this one should be better motivated. there are a lot of prev works exploring FT localized in a subspace, which we should connect to}
% \hao{suggest merging all three tales into one}
% \hao{it looks like this one will be significantly revised. i'm holding off a detailed pass}
% \textbf{Motivation and Research Question: }
This section aims to answer the following research question:
\begin{itemize}[noitemsep,topsep=0pt,parsep=0pt,partopsep=0pt]
\item[\textbf{RQ4:}] \textit{How consistent is the RL-updated subnetwork under varying training conditions such as random seed, training data, RL algorithm, and even all of them?
}
\end{itemize}

If the subnetwork remains largely consistent across these variations,
it would suggest that the identified subnetwork is not merely an artifact of specific training configuration but a generalizable and transferable structure of the pretrained model.

\noindent\textbf{Setup.}
To quantify the similarity between two subnetworks, we define an overlap metric. Let $s_1$ and $s_2$ denote the sparsity levels of two models, and let $\mathcal{I}_1$ and $\mathcal{I}_2$ be the sets of indices of the updated parameters. The size of the common subnetwork is given by $|\mathcal{I}_1 \cap \mathcal{I}_2|$.
One-sided overlap is then \(o_{1} = |\mathcal{I}_1 \cap \mathcal{I}_2|/|\mathcal{I}_1| = |\mathcal{I}_1 \cap \mathcal{I}_2|/(1-s_1)\), which quantifies the proportion of $\mathcal{I}_1$ that is covered by the subnetwork $\mathcal{I}_2$, i.e., how well $\mathcal{I}_2$ captures the parameters updated in $\mathcal{I}_1$.
Similarly, \(o_{2} = |\mathcal{I}_1 \cap \mathcal{I}_2|/(1-s_2)\) quantifies how well $\mathcal{I}_1$ captures the parameters updated in $\mathcal{I}_2$. We compare the observed overlaps $o_{1}$ and $o_2$ against a random guessing baseline,  where a subnetwork is constructed by uniformly selecting the same number of parameters as identified by RL (Appendix \ref{sec:random_guessing}). 
% We define the overlap as: \(\mathcal{O}_{1} = \frac{|\mathcal{I}_1 \cap \mathcal{I}_2|}{|\mathcal{I}_1|} = \frac{|\mathcal{I}_1 \cap \mathcal{I}_2|}{1-s_1}\) and \(\mathcal{O}_{2} = \frac{|\mathcal{I}_1 \cap \mathcal{I}_2|}{|\mathcal{I}_2|} = \frac{|\mathcal{I}_1 \cap \mathcal{I}_2|}{1-s_2}\)
% That is, we compute the number of shared non-zero gradient indices and normalize this count either by the total number of entries in the subnetwork of model 1 or in model 2. 

% As a baseline, we consider the expected one-sided overlap under random guessing. 
% \hao{i dropped this below. i think it is a better fit in the appendix}
% \hao{i'm too tired to check the math below}

% All experiments in this section are conducted using DPO, with variations introduced in random seeds, training data, and the choice of RL algorithm. 
We evaluate three settings: (1) varying the random seed alone, (2) varying the training data alone, and (3) changing the seed, data, and the RL algorithm altogether as a stress test.
We conduct controlled experiments and all factors not under investigation are the same.
% The changes in seed and training dataset are controlled experiments, with all other settings the same.
% We also conduct a stress test changing the RL algorithm, random seeds, and training data.
% In contrast, varying the RL algorithm serves as a stress test, introducing extensive changes where nearly all aspects of the training pipeline differ.
% For the random seed and the training data, we keep the training hyperparameters same, and only control for the dimension. 
Unless otherwise mentioned, all ablations were done with a batch size of 32, trained for one epoch with the base model \texttt{\href{https://huggingface.co/allenai/Llama-3.1-Tulu-3-8B-SFT}{Tulu-3-8B-SFT}}. 
% Further details are as follows:
% \textbf{Seed: } we used different random seeds, namely 42 and 123. 
When varying the training data, we switch between the Tulu preference dataset\footnote{\texttt{allenai/llama-3.1-tulu-3-8b-preference-mixture} } and the PRIME rollout dataset\footnote{ \texttt{PRIME-RL/Eurus-2-Rollout}. It has model generations for math datasets alongside a label for correctness}. 
To adapt the rollout dataset to \textsc{DPO} format, we select only positive samples, and pair it with a randomly sampled negative one. 
When varying the RL algorithm, we train a \textsc{DPO} model initialized from \texttt{PRIME-RL/Eurus-2-7B-SFT} and compare it to the \texttt{PRIME-RL/Eurus-2-7B-PRIME} model. 
% Although both models share the same initialization, their training paradigms differ substantially: \textsc{DPO} is offline and off-policy from an outcome-reward model, while \textsc{PRIME} is trained in an online and on-policy manner with process a reward model.

\noindent\textbf{Results.}
Table \ref{tab:merged_sparsity_overlap} reports our observed overlap.
% We compare against a random-guessing baseline, where a subnetwork is constructed by uniformly selecting the same number of parameters as identified by RL.
% For each model the column overlap indicates the overlap metric as defined earlier, where the normalization was done by the model's own subnetwork. 
% \noindent\textbf{Variation in Random Seed}
% Table \ref{tab:seed_sparsity_overlap} reports the observed overlap as compared to a random baseline. 
Despite changes in initialization, the resulting subnetworks show substantial overlap—well above the random baseline. For instance, varying the random seed yields overlaps of $o_1 = 60.5\%$ and $o_2 = 60.6\%$. Similar consistency is observed when the training dataset is varied. Even under a stress test, where 
the data, seed, and RL algorithm are all changed, we still observe notable overlaps of 59.1\% and 33.2\%. These findings indicate the presence of a subnetwork that is at least partially transferrable to other different settings.

\begin{table}[t]
    \centering
    \caption{Subnetwork overlap varying random seeds, training data, and RL algorithms. Despite these changes, subnetworks show non-trivial overlap compared to random-guessing baselines.}
    \setlength{\tabcolsep}{6pt}
    \renewcommand{\arraystretch}{1.1}
    \begin{tabular}{@{}lcccc@{}}
        \toprule
        \textbf{Variation Axis} & \textbf{Setting} & \textbf{Random} & \textbf{RL Subnetwork} & \textbf{Sparsity} \\
        \midrule
        \multirow{2}{*}{Seed} 
        & $\mathcal{I}_1$: 42  & $o_1$: 36.7 & $o_1$: 60.5 & 63.3 \\
        & $\mathcal{I}_2$: 123 & $o_2$: 36.7 & $o_2$: 60.6 & 63.3 \\
        \midrule
        \multirow{2}{*}{Data}
        & $\mathcal{I}_1$: Tulu Data & $o_1$: 14.6 & $o_1$: 26.7 & 63.3 \\
        & $\mathcal{I}_2$: PRIME Data    & $o_2$: 36.7 & $o_2$: 67.1 & 85.4 \\
        \midrule
        \multirow{2}{*}{Seed + Data + Algo.}
        & $\mathcal{I}_1$: DPO  & $o_1$: 23.0 & $o_1$: 59.1 & 87.1 \\
        & $\mathcal{I}_2$: PRIME & $o_2$: 12.9 & $o_2$: 33.2 & 77.0 \\
        \bottomrule
    \end{tabular}
    \label{tab:merged_sparsity_overlap}
    \vspace{-.5cm}
\end{table}

% \noindent\textbf{Variation in Training Dataset}
% For this experiment, we evaluate the effect of dataset distribution on the resulting dense subnetwork. Specifically, 

% It is important to note that these datasets are arbitrarily selected and are not expected to exhibit any     similarity in distribution. We report the sparsity of the independently trained models and compute the overlap between their dense subnetworks using the metric defined earlier. 

% \noindent\textbf{Variation in Algorithm}

\begin{AIbox}{Takeaway 3}
For a given base model, we observe substantially higher subnetwork overlap than random guessing across different seeds, training data, and RL algorithms. This suggests the potential of a consistent and at least partially transferrable subnetwork structure across these different training settings. %variations.
\end{AIbox}

% \hao{i added below. check:}
While the observed subnetwork overlap across seeds, datasets, and training algorithms falls short of 100\%, it suggests that partial subnetwork reuse may still offer practical utility. 
In particular, partial subnetwork reuse could reduce redundant computation across repeated RL runs, such as those in hyperparameter sweeps or ablation studies, by partially reusing the subnetworks. 
In addition, one might be able to reuse part of the subnetwork identified by a cheaper algorithm like DPO and reuse it in more expensive ones like PPO, greatly reducing the training cost without sacrificing performance.

% Our findings suggest that, despite these variations, the dense subnetwork for a given model remains largely similar as compared to a random baseline.
% Such consistency opens up exciting possibilities for reusable, transferable, and even precomputing subnetworks for downstream tasks. Prior work \citep{chen2020lotterytickethypothesispretrained} already talks about such use-cases for encoder models such as BERT \citep{devlin-etal-2019-bert} where the winning lottery tickets for the Masked Language Model task is used for downstream tasks without loss in performance. 
% A systematic exploration of such applications is beyond the scope of this work

\section{Why Do the Subnetworks Emerge?}
\label{sec:reason}
This section answers investigates the following research question:
\begin{itemize}[noitemsep,topsep=0pt,parsep=0pt,partopsep=0pt]
\item[\textbf{RQ5:}] \textit{What factors contribute to the update sparsity observed in RL finetuning?}\footnote{
We conjecture that, for the sake of illustration, if one were to perform backpropagation manually on paper with unlimited numerical precision, the resulting parameter updates would be dense. In practice, however, modern computers rely on floating-point arithmetic with limited precision. As a result, updates with very small magnitudes (e.g., absolute values below $10^{-40}$) cannot be represented and are effectively discarded, hence the empirically observed sparsity of the parameter updates. Importantly, such near-zero updates appear to have negligible impact on model performance, as strong results have been achieved without them. The equivalent question we address in this section is: What factors contribute to these near-zero parameter updates during RL fine-tuning?
}
\end{itemize}
\begin{wrapfigure}{r}{0.50\textwidth}
    \centering
    \vspace{-8pt}
    \includegraphics[width=0.50\textwidth]{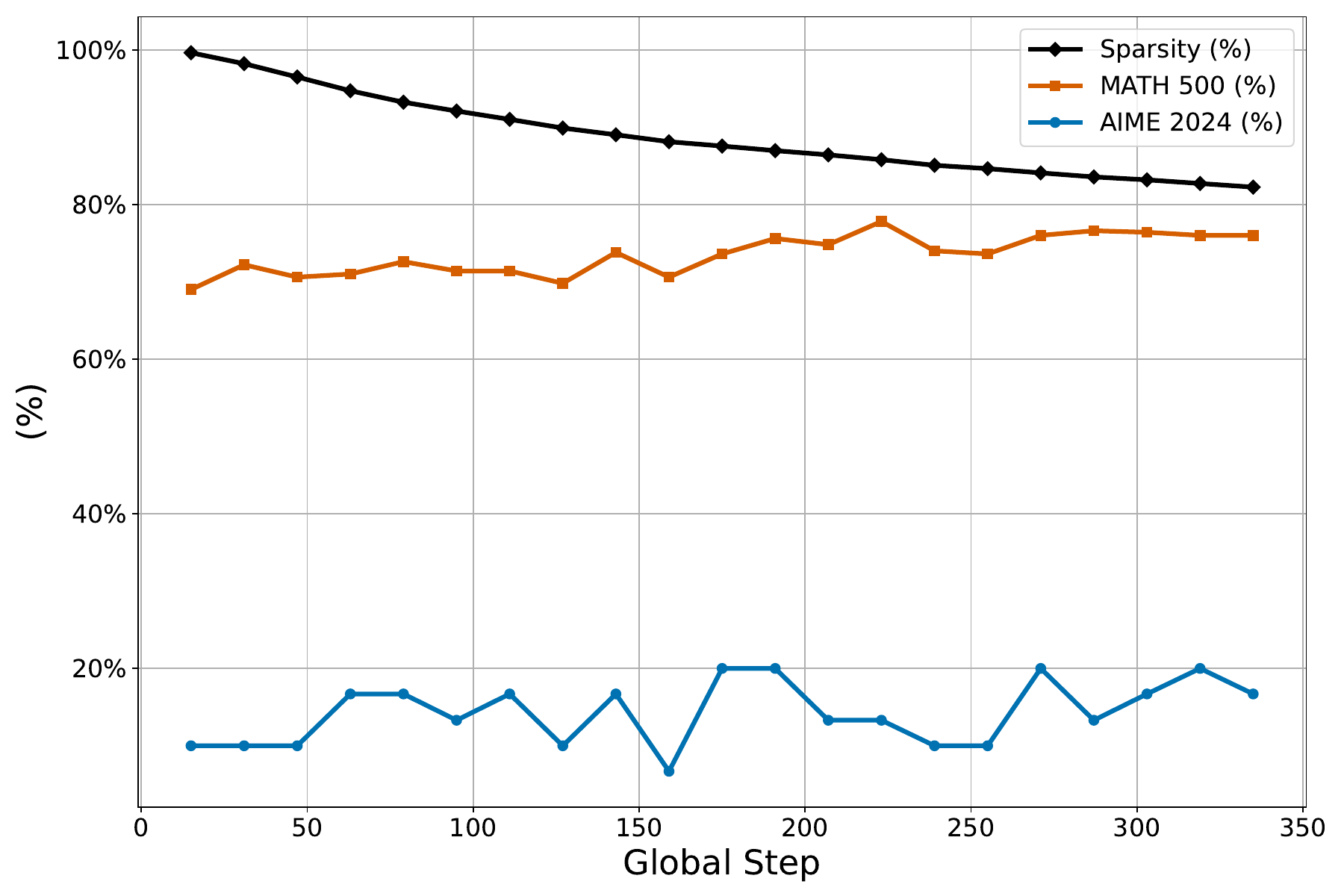}
    \vspace{-18pt}
    \caption{Update sparsity of intermediate checkpoints of a training run of PRIME. We observe that with more training the sparsity slowly decays.}
    \vspace{-10pt}
    \label{fig:intermediate_sparsity_prime}
\end{wrapfigure} 
% \textbf{RQ5:} \textit

% To better understand the origins of sparsity in RL-induced parameter updates, 
We investigate the following factors: gradient clipping, KL regularization towards a reference policy, performing SFT prior to RL, and the number of RL update steps. Our investigation suggests that  the dominating factor is how close the training data distribution is to the policy's, i.e., whether the training data is in-distribution.
Another  important factor is the total number of gradient updates.
% We now investigate the root causes behind the sparsity observed. Specifically, we ask: is this sparsity driven by gradient clipping, KL divergence, normalization against the base policy, the use of supervised fine-tuning, or simply the number of RL updates? Notably, for most of the potential root causes, we negate them with at least one strong negative evidence. 
% Our empirical findings suggest that the \textbf{dominant factor} is the \textit{on-policyness of the data.}. Another crucial factor is the number of gradient steps performed while training. 

% \hao{we need to explain why these factors can affect sparsity}

\noindent\textbf{Gradient clipping and KL regularization.} 
As discussed in \S\ref{sec:background}, gradient clipping and KL regularization are commonly used to keep the policy from deviating too far away from a reference model.
Since both mechanisms explicitly suppress large parameter updates, they are natural candidates for contributing factors to the observed update sparsity.
To test their impact, we train a GRPO variant using Qwen-2.5-7B-Instruct, comparing models with and without these regularization terms. We find that both configurations exhibit comparable sparsity levels, suggesting that neither gradient clipping nor KL regularization is a primary driver of the update sparsity. In our experiments, the GRPO variant trained with KL regularization achieved a sparsity of 69.8\%, while the variant trained without KL regularization reached 68.8\%. Further, SimPO removes the KL term by dropping the base policy normalization in DPO, and as reported in \ref{tab:gross_sparsity} SimPO also produces sparse updates, providing further negative evidence for KL.
% \lifan{where is the result?}

% To test the influence of gradient clipping and KL regularization, we train a GRPO variant using \texttt{Qwen-2.5-7B-Instruct}. We compare checkpoints with and without these terms and find both exhibit comparable sparsity levels, indicating that neither is a decisive factor.
\begin{wrapfigure}{r}{0.52\textwidth}
    \centering
    \vspace{-10pt}
    \includegraphics[width=0.52\textwidth]{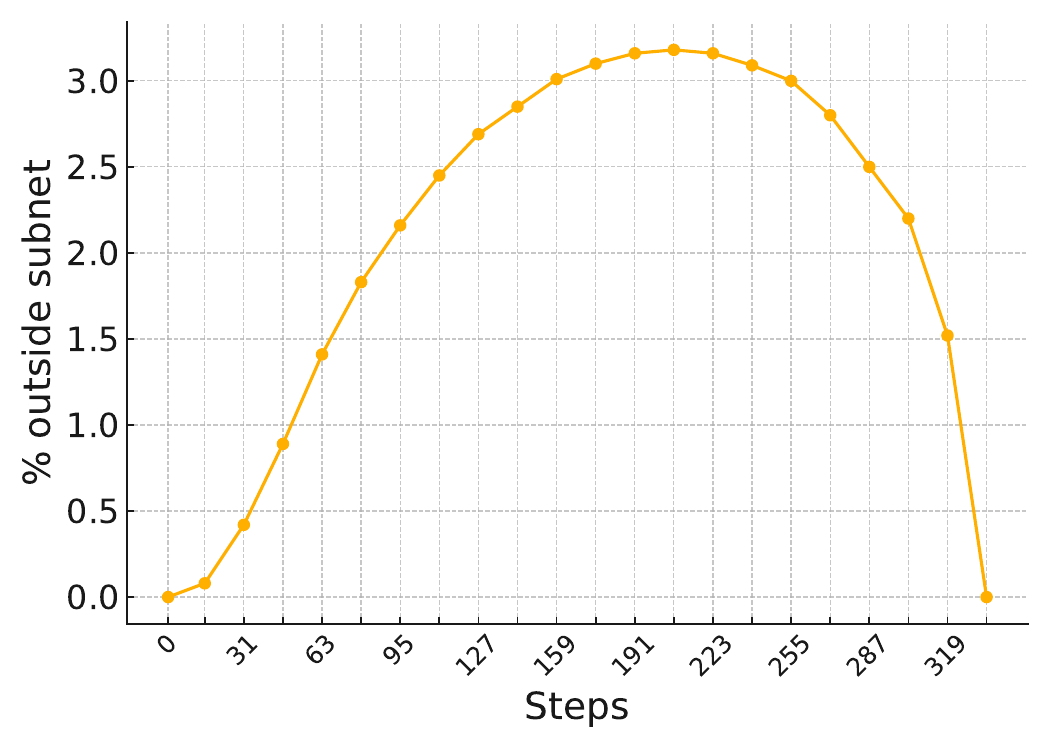}
    \vspace{-10pt}
    \caption{ Percentage of updated weights that are outside the final subnetwork across training steps}% note that this number steadily increases then decreases indicating that updates cancel each other out. }
    \vspace{-15pt}
    \label{fig:subnetwork_cancellation}
\end{wrapfigure}
\noindent\textbf{Performing SFT before RL.} 
A common design choice is to perform SFT on the same data as the subsequent RL \cite{ouyang2022traininglanguagemodelsfollow}.
However, as shown in Table~\ref{tab:gross_sparsity}, our findings extend to models such as DeepSeek-R1-Zero, which forgoes SFT entirely yet still exhibit high update sparsity. 
This suggests that  SFT is \emph{not} a main contributing factor to the update sparsity in RL finetuning.
% \lifan{i remember we previously talked about simpo.also worth mentioning if space permits}
% Another popular design choice (often ignored in more recent works) is performing SFT on the target datasets before the RL training initiates. However as observed in Table \ref{tab:gross_sparsity}, our observations extend to models like \texttt{DeepSeek-R1-Zero}, which omit the SFT stage entirely, providing negative evidence for the same. 

% \textbf{Base Policy Normalization:} 
% Another potential design choice in RL is the normalization with the base policy. Our observations generalize to models like SimPO, which omit the normalization term entirely. These still exhibit high sparsity (as reported in Table \ref{tab:gross_sparsity}), suggesting that normalization against the base policy is not the root cause.

\noindent\textbf{Training duration.} 
It is intuitive that with more gradient steps, a model is expected to drift more from the base model. 
Figure~\ref{fig:intermediate_sparsity_prime} shows how update sparsity changes during the  training of PRIME.
As training progresses, update sparsity gradually decreases but eventually shows signs of convergence to around 80\%.
Notably, \texttt{DeepSeek-R1-Zero} undergoes 8K training steps (numbers from Figure 2 in \cite{deepseekai2025deepseekr1incentivizingreasoningcapability}) using GRPO, over 20$\times$ more than PRIME, but shows a comparable update sparsity (86\%). Therefore, we conjecture that training duration's impact on update sparsity is more prominent during early training but gradually decreases as training progresses.

Figure~\ref{fig:subnetwork_cancellation} shows the percentage of updated parameters (relative to model size) that lie outside the final subnetwork. This proportion increases during the early stages of training but steadily declines in later stages. This trend suggests that some parameters outside the final subnetwork receive non-zero gradient updates that cancel out.
Overall, about 8.5\% of parameters that are ever updated during training fall outside the final subnetwork.

While it is possible, though less likely, that all models we study, including \texttt{DeepSeek-R1-Zero} (86.0\%  update sparsity after 8K steps),
are severely undertrained, and that the observed update sparsity would diminish with substantially more training.
Nonetheless, we question the practicality of this hypothesis since it runs counter to the RL literature arguing against overtraining to 
prevent overfitting and improve generalization \cite{pmlr-v97-fu19a}.

% \hao{we should clarify that the checkpoints are trained in a realistic way maximizing validation accuracy instead of cherried picked to show high sparsity. we should also cite evidence against over training in rl}
% \hao{we need to actually say something about the table. }
% \hao{what 's a realistic \# steps? is the 8K one significantly overtraining just for illustration purposes? does it hurt the performance? we need to be careful about the wordings here since otherwise the finding might look trivial to some}
% \hao{do we have a control? say, sft with the same amount of updates?}
% \hao{see slack discussion for suggestions}
% \lifan{must-do: emphasize deepseek-r1-zero. it has gone through over 8k steps, the most among the model tested, but the sparsity still holds. otherwise one may read our results as undertrainning.}

\looseness=-1
\noindent\textbf{Training on in-distribution data.}
% \lifan{the setup is a bit messy. i still don't know how many comparisons we have here.}
% We conjecture that the update sparsity
% emerges from trainining on in‑distribution data. 
Intuitively, when gradients are computed on sequences that the policy already assigns high probabilities to,  little update to the parameters would be needed.
We evaluated two scenarios: (1) rejection sampling, and
(2) DPO on out-of-distribution data by \emph{not} performing SFT prior to RL.
Since we've already learned from \S\ref{sec:empirical_evidence} that 
DPO with in-distribution data induces sparse updates while SFT (with out-of-distribution data) induces dense ones, this additional experiment can serve as a control group to isolate  the factor of training on in- vs. out-of-distribution data.

As shown in Table~\ref{tab:sparsity_rejection}, Our experiments reveal that SFT on in-distribution data produces sparse updates, while DPO with out-of-distribution data produces dense ones.
Specifically,  performing SFT with  \texttt{Qwen/Qwen2.5-Math-7B}  on rejection sampled in-distribution data yields around 90.0\% update sparsity. 
This is reinforced by the examination of a previous work: RAFT\texttt{++} \citep{xiong2024iterative}, which performs supervised finetuning with iterative rejection sampling,  
yields an update sparsity of 69.4\%. 
In contrast, DPO on out-of-distribution data produces dense updates in \href{https://huggingface.co/HuggingFaceH4/zephyr-7b-beta}{\texttt{zephyr-7b-beta}} models, with a 6.8\% update sparsity. 
These findings suggest that training on in-distribution data could be a major  driver of update sparsity in not only RL, but also SFT.

\begin{AIbox}{Takeaway 4}
We conjecture that training on in-distribution data could be a reason of update sparsity;
KL-divergence regularization and gradient clipping have limited impact.
\end{AIbox}

\begin{table}[h]
% \small
\centering
\caption{RFT indicates rejection‐sampling fine‑tuning \citep{touvron2023llama2openfoundation,dong2023raftrewardrankedfinetuning}, and RAFT++ is iterative RFT  A comparative analysis across SFT and DPO as well as in- vs out-of-distribution training shows that in-distribution consistently produces sparse updates.}
\setlength{\tabcolsep}{8pt}
\renewcommand{\arraystretch}{1.2}
\begin{tabular}{lccccc}
\toprule
\textbf{Model} & \textbf{Method} & \textbf{Sparsity (\%)} & \textbf{SFT/RL} & \textbf{In‐Dist} \\
\midrule
\texttt{Qwen2.5-Math-7B} & RFT  & 91.2 & SFT & \cmark \\
\texttt{Qwen2.5-Math-7B} & RAFT++                  & 69.4 & SFT & \cmark\\
\texttt{Llama-3.1-8B-SFT}  & SFT                     & 6.8 & SFT & \xmark \\
\texttt{Llama-3.1-8B-SFT}  & DPO                     & 6.8 & RL & \xmark \\
\texttt{Zephyr-7b-Beta}  & DPO                     & 7.7 & RL & \xmark \\
\texttt{Llama-3.1-8B-DPO}  & DPO                     & 81.4 & RL & \cmark \\
\bottomrule
\end{tabular}
\label{tab:sparsity_rejection}
\vspace{-.25cm}
\end{table}

\section{Limitations and Future Work}
\label{sec:limitations}
% Our study reveals that RL fine‑tuning of LLMs produces highly sparse parameter updates. 
Because RL is computationally demanding, we choose to vary one factor at a time; yet the observed sparsity may actually result from complex interactions among many, an avenue future work should examine. 
Further, fully controlled experiments are computationally prohibitive, and we thus sometimes resort to resort to public checkpoints. 
While our experiments focus on language models, it would be interesting to explore the same questions for multimodal and diffusion models. 
Subsequent research could investigate methods for early identification of the sparse subnetwork and ways to leverage its structure for more efficient learning. 
Finally, our empirical findings invite a deeper theoretical analysis, with the goal of uncovering theoretical explanations for the update sparsity in RL. Lastly, while our observations generally hold, there are confounders (for eg. \cite{liu2025prorlprolongedreinforcementlearning}). However the confounding reason is not trivial and is worth exploring as part of future work. Further to the best of our observations this phenomenon is largely applicable to all RL settings. 

\section{Conclusion}
\label{sec:conclusion}
Our study reveals that RL finetuning in LLMs  updates only a sparse subnetwork constituting approximately 5\%-30\% of total parameters, leaving the rest unchanged. 
This sparsity emerges without explicit spasity promoting techniques such as regularization or structural constraints. Crucially, finetuning just this subnetwork in isolation reproduces the full model’s performance, aligning closely with original parameter values. For a given base model across different seeds, datasets and learning algorithms, a non-trivial portion of the subnetwork remains the same. Our findings highlight that learning from in-distribution samples while training is a key driver of this phenomenon, pointing towards more efficient and effective training strategies in RL-based finetuning of LLMs.
\section{Acknowledgments}
We would like to thank Pavan Jayasinha, Cheng Wang, Abdul Waheed, Shivanshu Shekhar, Alexander Schwing, Gabriel Stanovsky, Roy Schwartz and Gokhan Tur for their valuable feedbacks and suggestions. Further we extend our gratitude to the members of ConvAI and Alta lab groups at UIUC for their thoughtful remarks on our draft. This research used the Delta advanced computing and data resource which is supported by the National Science Foundation (award OAC 2005572) and the State of Illinois. Delta is a joint effort of the University of Illinois Urbana-Champaign and its National Center for Supercomputing Applications.
% This suggests potential for substantial computational savings and scalability in practical applications.

\bibliography{custom}
\bibliographystyle{plainnat}
\clearpage

\section*{NeurIPS Paper Checklist}

%%% BEGIN INSTRUCTIONS %%%
The checklist is designed to encourage best practices for responsible machine learning research, addressing issues of reproducibility, transparency, research ethics, and societal impact. Do not remove the checklist: {\bf The papers not including the checklist will be desk rejected.} The checklist should follow the references and follow the (optional) supplemental material.  The checklist does NOT count towards the page
limit. 

Please read the checklist guidelines carefully for information on how to answer these questions. For each question in the checklist:
\begin{itemize}
    \item You should answer \answerYes{}, \answerNo{}, or \answerNA{}.
    \item \answerNA{} means either that the question is Not Applicable for that particular paper or the relevant information is Not Available.
    \item Please provide a short (1–2 sentence) justification right after your answer (even for NA). 
   % \item {\bf The papers not including the checklist will be desk rejected.}
\end{itemize}

{\bf The checklist answers are an integral part of your paper submission.} They are visible to the reviewers, area chairs, senior area chairs, and ethics reviewers. You will be asked to also include it (after eventual revisions) with the final version of your paper, and its final version will be published with the paper.

The reviewers of your paper will be asked to use the checklist as one of the factors in their evaluation. While "\answerYes{}" is generally preferable to "\answerNo{}", it is perfectly acceptable to answer "\answerNo{}" provided a proper justification is given (e.g., "error bars are not reported because it would be too computationally expensive" or "we were unable to find the license for the dataset we used"). In general, answering "\answerNo{}" or "\answerNA{}" is not grounds for rejection. While the questions are phrased in a binary way, we acknowledge that the true answer is often more nuanced, so please just use your best judgment and write a justification to elaborate. All supporting evidence can appear either in the main paper or the supplemental material, provided in appendix. If you answer \answerYes{} to a question, in the justification please point to the section(s) where related material for the question can be found.

% IMPORTANT, please:
% \begin{itemize}
%     \item {\bf Delete this instruction block, but keep the section heading ``NeurIPS Paper Checklist"},
%     \item  {\bf Keep the checklist subsection headings, questions/answers and guidelines below.}
%     \item {\bf Do not modify the questions and only use the provided macros for your answers}.
% \end{itemize} 

%%% END INSTRUCTIONS %%%

\begin{enumerate}

\item {\bf Claims}
    \item[] Question: Do the main claims made in the abstract and introduction accurately reflect the paper's contributions and scope?
    \item[] Answer: \answerYes{} % Replace by \answerYes{}, \answerNo{}, or \answerNA{}.
    \item[] Justification: Sections 3-6 provides empirical evidence for the claims made in intro and abstract
    % \item[] Guidelines:
    % \begin{itemize}
    %     \item The answer NA means that the abstract and introduction do not include the claims made in the paper.
    %     \item The abstract and/or introduction should clearly state the claims made, including the contributions made in the paper and important assumptions and limitations. A No or NA answer to this question will not be perceived well by the reviewers. 
    %     \item The claims made should match theoretical and experimental results, and reflect how much the results can be expected to generalize to other settings. 
    %     \item It is fine to include aspirational goals as motivation as long as it is clear that these goals are not attained by the paper. 
    % \end{itemize}

\item {\bf Limitations}
    \item[] Question: Does the paper discuss the limitations of the work performed by the authors?
    \item[] Answer: \answerYes{} % Replace by \answerYes{}, \answerNo{}, or \answerNA{}.
    \item[] Justification: Separate dedicated limiation section included

\item {\bf Theory assumptions and proofs}
    \item[] Question: For each theoretical result, does the paper provide the full set of assumptions and a complete (and correct) proof?
    \item[] Answer: \answerNA{} % Replace by \answerYes{}, \answerNo{}, or \answerNA{}.
    \item[] Justification: The paper doesnt discuss theoretical proofs, most results are empirical
    % \item[] Guidelines:
    % \begin{itemize}
    %     \item The answer NA means that the paper does not include theoretical results. 
    %     \item All the theorems, formulas, and proofs in the paper should be numbered and cross-referenced.
    %     \item All assumptions should be clearly stated or referenced in the statement of any theorems.
    %     \item The proofs can either appear in the main paper or the supplemental material, but if they appear in the supplemental material, the authors are encouraged to provide a short proof sketch to provide intuition. 
    %     \item Inversely, any informal proof provided in the core of the paper should be complemented by formal proofs provided in appendix or supplemental material.
    %     \item Theorems and Lemmas that the proof relies upon should be properly referenced. 
    % \end{itemize}

    \item {\bf Experimental result reproducibility}
    \item[] Question: Does the paper fully disclose all the information needed to reproduce the main experimental results of the paper to the extent that it affects the main claims and/or conclusions of the paper (regardless of whether the code and data are provided or not)?
    \item[] Answer: \answerYes{} % Replace by \answerYes{}, \answerNo{}, or \answerNA{}.
    \item[] Justification: All relevant datasets, models are listed, all hyperparameters discussed in detail

\item {\bf Open access to data and code}
    \item[] Question: Does the paper provide open access to the data and code, with sufficient instructions to faithfully reproduce the main experimental results, as described in supplemental material?
    \item[] Answer: \answerYes{} % Replace by \answerYes{}, \answerNo{}, or \answerNA{}.
    \item[] Justification: 
    \item[] Guidelines:
    % \begin{itemize}
    %     \item The answer NA means that paper does not include experiments requiring code.
    %     \item Please see the NeurIPS code and data submission guidelines (\url{https://nips.cc/public/guides/CodeSubmissionPolicy}) for more details.
    %     \item While we encourage the release of code and data, we understand that this might not be possible, so “No” is an acceptable answer. Papers cannot be rejected simply for not including code, unless this is central to the contribution (e.g., for a new open-source benchmark).
    %     \item The instructions should contain the exact command and environment needed to run to reproduce the results. See the NeurIPS code and data submission guidelines (\url{https://nips.cc/public/guides/CodeSubmissionPolicy}) for more details.
    %     \item The authors should provide instructions on data access and preparation, including how to access the raw data, preprocessed data, intermediate data, and generated data, etc.
    %     \item The authors should provide scripts to reproduce all experimental results for the new proposed method and baselines. If only a subset of experiments are reproducible, they should state which ones are omitted from the script and why.
    %     \item At submission time, to preserve anonymity, the authors should release anonymized versions (if applicable).
    %     \item Providing as much information as possible in supplemental material (appended to the paper) is recommended, but including URLs to data and code is permitted.
    % \end{itemize}

\item {\bf Experimental setting/details}
    \item[] Question: Does the paper specify all the training and test details (e.g., data splits, hyperparameters, how they were chosen, type of optimizer, etc.) necessary to understand the results?
    \item[] Answer: \answerYes{} % Replace by \answerYes{}, \answerNo{}, or \answerNA{}.
    \item[] Justification: All details shared for reproducibility
    \item[] Guidelines:
    % \begin{itemize}
    %     \item The answer NA means that the paper does not include experiments.
    %     \item The experimental setting should be presented in the core of the paper to a level of detail that is necessary to appreciate the results and make sense of them.
    %     \item The full details can be provided either with the code, in appendix, or as supplemental material.
    % \end{itemize}

\item {\bf Experiment statistical significance}
    \item[] Question: Does the paper report error bars suitably and correctly defined or other appropriate information about the statistical significance of the experiments?
    \item[] Answer: \answerNo{}{} % Replace by \answerYes{}, \answerNo{}, or \answerNA{}.
    \item[] Justification: Most accuracies are reported on deterministic setups
    % \item[] Guidelines:
    % \begin{itemize}
    %     \item The answer NA means that the paper does not include experiments.
    %     \item The authors should answer "Yes" if the results are accompanied by error bars, confidence intervals, or statistical significance tests, at least for the experiments that support the main claims of the paper.
    %     \item The factors of variability that the error bars are capturing should be clearly stated (for example, train/test split, initialization, random drawing of some parameter, or overall run with given experimental conditions).
    %     \item The method for calculating the error bars should be explained (closed form formula, call to a library function, bootstrap, etc.)
    %     \item The assumptions made should be given (e.g., Normally distributed errors).
    %     \item It should be clear whether the error bar is the standard deviation or the standard error of the mean.
    %     \item It is OK to report 1-sigma error bars, but one should state it. The authors should preferably report a 2-sigma error bar than state that they have a 96\% CI, if the hypothesis of Normality of errors is not verified.
    %     \item For asymmetric distributions, the authors should be careful not to show in tables or figures symmetric error bars that would yield results that are out of range (e.g. negative error rates).
    %     \item If error bars are reported in tables or plots, The authors should explain in the text how they were calculated and reference the corresponding figures or tables in the text.
    % \end{itemize}

\item {\bf Experiments compute resources}
    \item[] Question: For each experiment, does the paper provide sufficient information on the computer resources (type of compute workers, memory, time of execution) needed to reproduce the experiments?
    \item[] Answer: \answerYes{} % Replace by \answerYes{}, \answerNo{}, or \answerNA{}.
    \item[] Justification: \justificationTODO{}
    % \item[] Guidelines:
    % \begin{itemize}
    %     \item The answer NA means that the paper does not include experiments.
    %     \item The paper should indicate the type of compute workers CPU or GPU, internal cluster, or cloud provider, including relevant memory and storage.
    %     \item The paper should provide the amount of compute required for each of the individual experimental runs as well as estimate the total compute. 
    %     \item The paper should disclose whether the full research project required more compute than the experiments reported in the paper (e.g., preliminary or failed experiments that didn't make it into the paper). 
    % \end{itemize}
    
\item {\bf Code of ethics}
    \item[] Question: Does the research conducted in the paper conform, in every respect, with the NeurIPS Code of Ethics \url{https://neurips.cc/public/EthicsGuidelines}?
    \item[] Answer: \answerYes{} % Replace by \answerYes{}, \answerNo{}, or \answerNA{}.
    \item[] Justification: Most experiments performed are with publicly released models, datasets and codebases.
    % \item[] Guidelines:
    % \begin{itemize}
    %     \item The answer NA means that the authors have not reviewed the NeurIPS Code of Ethics.
    %     \item If the authors answer No, they should explain the special circumstances that require a deviation from the Code of Ethics.
    %     \item The authors should make sure to preserve anonymity (e.g., if there is a special consideration due to laws or regulations in their jurisdiction).
    % \end{itemize}

\item {\bf Broader impacts}
    \item[] Question: Does the paper discuss both potential positive societal impacts and negative societal impacts of the work performed?
    \item[] Answer: \answerNA{} % Replace by \answerYes{}, \answerNo{}, or \answerNA{}.
    \item[] Justification: No immediate societal impacts of the work

\item {\bf Safeguards}
    \item[] Question: Does the paper describe safeguards that have been put in place for responsible release of data or models that have a high risk for misuse (e.g., pretrained language models, image generators, or scraped datasets)?
    \item[] Answer: \answerNA{} % Replace by \answerYes{}, \answerNo{}, or \answerNA{}.
    \item[] Justification: No such risks
    % \item[] Guidelines:
    % \begin{itemize}
    %     \item The answer NA means that the paper poses no such risks.
    %     \item Released models that have a high risk for misuse or dual-use should be released with necessary safeguards to allow for controlled use of the model, for example by requiring that users adhere to usage guidelines or restrictions to access the model or implementing safety filters. 
    %     \item Datasets that have been scraped from the Internet could pose safety risks. The authors should describe how they avoided releasing unsafe images.
    %     \item We recognize that providing effective safeguards is challenging, and many papers do not require this, but we encourage authors to take this into account and make a best faith effort.
    % \end{itemize}

\item {\bf Licenses for existing assets}
    \item[] Question: Are the creators or original owners of assets (e.g., code, data, models), used in the paper, properly credited and are the license and terms of use explicitly mentioned and properly respected?
    \item[] Answer: \answerYes{} % Replace by \answerYes{}, \answerNo{}, or \answerNA{}.
    \item[] Justification: All works are cited.
    % \item[] Guidelines:
    % \begin{itemize}
    %     \item The answer NA means that the paper does not use existing assets.
    %     \item The authors should cite the original paper that produced the code package or dataset.
    %     \item The authors should state which version of the asset is used and, if possible, include a URL.
    %     \item The name of the license (e.g., CC-BY 4.0) should be included for each asset.
    %     \item For scraped data from a particular source (e.g., website), the copyright and terms of service of that source should be provided.
    %     \item If assets are released, the license, copyright information, and terms of use in the package should be provided. For popular datasets, \url{paperswithcode.com/datasets} has curated licenses for some datasets. Their licensing guide can help determine the license of a dataset.
    %     \item For existing datasets that are re-packaged, both the original license and the license of the derived asset (if it has changed) should be provided.
    %     \item If this information is not available online, the authors are encouraged to reach out to the asset's creators.
    % \end{itemize}

\item {\bf New assets}
    \item[] Question: Are new assets introduced in the paper well documented and is the documentation provided alongside the assets?
    \item[] Answer: \answerNA{} % Replace by \answerYes{}, \answerNo{}, or \answerNA{}.
    \item[] Justification: Doesnt release new assets
    % \item[] Guidelines:
    % \begin{itemize}
    %     \item The answer NA means that the paper does not release new assets.
    %     \item Researchers should communicate the details of the dataset/code/model as part of their submissions via structured templates. This includes details about training, license, limitations, etc. 
    %     \item The paper should discuss whether and how consent was obtained from people whose asset is used.
    %     \item At submission time, remember to anonymize your assets (if applicable). You can either create an anonymized URL or include an anonymized zip file.
    % \end{itemize}

\item {\bf Crowdsourcing and research with human subjects}
    \item[] Question: For crowdsourcing experiments and research with human subjects, does the paper include the full text of instructions given to participants and screenshots, if applicable, as well as details about compensation (if any)? 
    \item[] Answer: \answerNA{} % Replace by \answerYes{}, \answerNo{}, or \answerNA{}.
    \item[] Justification: Doesn't involve crowdsourcing. 
    % \item[] Guidelines:
    % \begin{itemize}
    %     \item The answer NA means that the paper does not involve crowdsourcing nor research with human subjects.
    %     \item Including this information in the supplemental material is fine, but if the main contribution of the paper involves human subjects, then as much detail as possible should be included in the main paper. 
    %     \item According to the NeurIPS Code of Ethics, workers involved in data collection, curation, or other labor should be paid at least the minimum wage in the country of the data collector. 
    % \end{itemize}

\item {\bf Institutional review board (IRB) approvals or equivalent for research with human subjects}
    \item[] Question: Does the paper describe potential risks incurred by study participants, whether such risks were disclosed to the subjects, and whether Institutional Review Board (IRB) approvals (or an equivalent approval/review based on the requirements of your country or institution) were obtained?
    \item[] Answer: \answerNA{} % Replace by \answerYes{}, \answerNo{}, or \answerNA{}.
    \item[] Justification: does not involve crowdsourcing nor research with human subjects
    % \item[] Guidelines:
    % \begin{itemize}
    %     \item The answer NA means that the paper does not involve crowdsourcing nor research with human subjects.
    %     \item Depending on the country in which research is conducted, IRB approval (or equivalent) may be required for any human subjects research. If you obtained IRB approval, you should clearly state this in the paper. 
    %     \item We recognize that the procedures for this may vary significantly between institutions and locations, and we expect authors to adhere to the NeurIPS Code of Ethics and the guidelines for their institution. 
    %     \item For initial submissions, do not include any information that would break anonymity (if applicable), such as the institution conducting the review.
    % \end{itemize}

\item {\bf Declaration of LLM usage}
    \item[] Question: Does the paper describe the usage of LLMs if it is an important, original, or non-standard component of the core methods in this research? Note that if the LLM is used only for writing, editing, or formatting purposes and does not impact the core methodology, scientific rigorousness, or originality of the research, declaration is not required.
    %this research? 
    \item[] Answer: \answerNA{} % Replace by \answerYes{}, \answerNo{}, or \answerNA{}.
    \item[] Justification: Does not involve LLMs usage
    % \item[] Guidelines:
    % \begin{itemize}
    %     \item The answer NA means that the core method development in this research does not involve LLMs as any important, original, or non-standard components.
    %     \item Please refer to our LLM policy (\url{https://neurips.cc/Conferences/2025/LLM}) for what should or should not be described.
    % \end{itemize}

\end{enumerate}

\newpage
\appendix
\section{Appendix}
\begin{table}[h]
\centering
\small
\begin{tabular}{llcccc}
\toprule
\textbf{Algorithm} & \textbf{Model (RL Checkpoint)} & \textbf{Tol = 1e-8} & \textbf{Tol = 1e-7} & \textbf{Tol = 1e-6} & \textbf{Tol = 1e-5} \\
\midrule
DPO   & allenai/Llama-3.1-Tulu-3-8B-DPO         & 76.04 & 76.04 & 76.14 & 81.38 \\
DPO   & allenai/Llama-3.1-Tulu-3-70B-DPO        & 87.58 & 87.59 & 87.79 & 95.24 \\
GRPO  & deepseek-ai/deepseek-math-7b-rl         & 68.14 & 68.14 & 68.14 & 68.53 \\
ORPO  & kaist-ai/mistral-orpo-beta              & 73.16 & 73.18 & 73.23 & 76.94 \\
ORPO  & kaist-ai/mistral-orpo-alpha             & 50.40 & 50.41 & 50.48 & 53.23 \\
KTO   & openbmb/Eurus-7b-kto                    & 71.78 & 71.79 & 73.14 & 95.98 \\
PPO   & peiyi9979/math-shepherd-mistral-7b-rl   & 52.45 & 52.47 & 53.21 & 80.77 \\
PPO   & PRIME-RL/Eurus-2-7B-PRIME               & 75.26 & 75.27 & 75.36 & 77.04 \\
SimPO & Llama-3-Instruct-8B-SimPO               & 71.00 & 71.00 & 71.10 & 76.42 \\
SimPO & Llama-3-Base-8B-SFT-SimPO               & 79.47 & 79.47 & 79.60 & 86.52 \\
SimPO & Mistral-7B-Instruct-SimPO               & 59.37 & 59.40 & 60.31 & 89.07 \\
SimPO & Mistral-7B-Base-SFT-SimPO               & 62.58 & 62.60 & 63.56 & 91.44 \\
\bottomrule
\end{tabular}
\caption{Sparsity (\%) of parameter updates under different thresholds across RL algorithms and RL checkpoints.}
\label{tab:full_sparsity_table}
\end{table}

\section{Hyperparameter choices for Gradient Masking experiments}
\label{sec:hyperparam_gradient_masking}
\textbf{DPO: } For DPO, we fine-tuned the \texttt{LLaMA-3.1-Tulu-3-8B} model using Direct Preference Optimization (DPO) with \texttt{bfloat16} mixed-precision and DeepSpeed Stage 3 for memory and compute efficiency across 8 processes. Training uses a sequence length of 2048 tokens with an effective batch size of 128, achieved by setting the per-device batch size to 1 with 16 gradient accumulation steps. A linear learning rate schedule is applied with a peak learning rate of $5 \times 10^{-7}$ and a warmup ratio of 0.1, without weight decay. The model is trained for one epoch on the \texttt{allenai/llama-3.1-tulu-3-8b-preference-mixture} dataset.

\textbf{PRIME: } For PRIME, We fine-tune \texttt{Qwen2.5-Math-7B} using on a mixture of GSM8K and MATH datasets. The training batch size is set to 64. The actor is optimized with a learning rate of $5 \times 10^{-7}$ while the reward model is trained with a learning rate of $1 \times 10^{-6}$. We performed four rollouts are performed per sample. We use gradient clipping of 10.0, and a temperature $\beta$ of 0.05. Training is conducted on for 15 epochs.

\section{Model Checkpoints: SFT vs RL sparsity comparison}
\label{sec:SFTvsRL_checkpoints}
\paragraph{SFT Checkpoints.}
We compare the following base and SFT checkpoints:
\begin{itemize}
\item \texttt{meta-llama/Llama-3.1-8B} vs. \texttt{allenai/Llama-3.1-Tulu-3-8B-SFT}
\item \texttt{meta-llama/Llama-3.1-70B} vs. \texttt{allenai/Llama-3.1-Tulu-3-70B-SFT}
\item \texttt{Qwen/Qwen2.5-Math-7B} vs. \texttt{PRIME-RL/Eurus-2-7B-SFT}
\end{itemize}
\paragraph{RL Checkpoints.}
We compare the following SFT and RL-finetuned checkpoints:
\begin{itemize}
\item \texttt{allenai/Llama-3.1-Tulu-3-8B-SFT} vs. \texttt{allenai/Llama-3.1-Tulu-3-8B-DPO}
\item \texttt{allenai/Llama-3.1-Tulu-3-70B-SFT} vs. \texttt{allenai/Llama-3.1-Tulu-3-70B-DPO}
\item \texttt{PRIME-RL/Eurus-2-7B-SFT} vs. \texttt{PRIME-RL/Eurus-2-7B-PRIME}
\end{itemize}

\section{Training Dynamics}
\label{sec:training_dynamics}
We analyzed intermediate checkpoints of the \textsc{Prime} model. Notably, our goal is to observe the convergence of the sparsity with training time. Does the sparsity decay with gradient steps ? Does it asymptotically reach a sparsity level of zero or is the convergence to a non-zero point ? 
\paragraph{Experimental setup}
We analyze 21 intermediate checkpoints from a training run of the PRIME model. Let the model parameters at these checkpoints be denoted by \(\theta_1, \theta_2, \ldots, \theta_{21}\), and let \(\theta_{\text{init}}\) denote the parameters of the corresponding base model (i.e., \texttt{PRIME-RL/Eurus-7b-sft}). We define the sparsity between the base model to checkpoint \(k\) as $sparsity_k$ = sparsity($\theta_k, \theta_{init}$), and the sparsity between two checkpoints \(i\) and \(j\) as $sparsity_{ij}$ = sparsity($\theta_j, \theta_i$).

\paragraph{Key Findings}
Our analysis begins by examining how the $sparsity_k$ evolve with training progress, offering insight into the update patterns.

\begin{figure}[h]
    \centering
    \begin{subfigure}[t]{0.48\textwidth}
        \centering
        \includegraphics[width=\textwidth]{eurus_sparsity_accuracy_plot_contrasting_tall_large_fonts.pdf}
        \caption{}
    \end{subfigure}
    \hfill
    \begin{subfigure}[t]{0.48\textwidth}
        \centering
        \includegraphics[width=\textwidth]{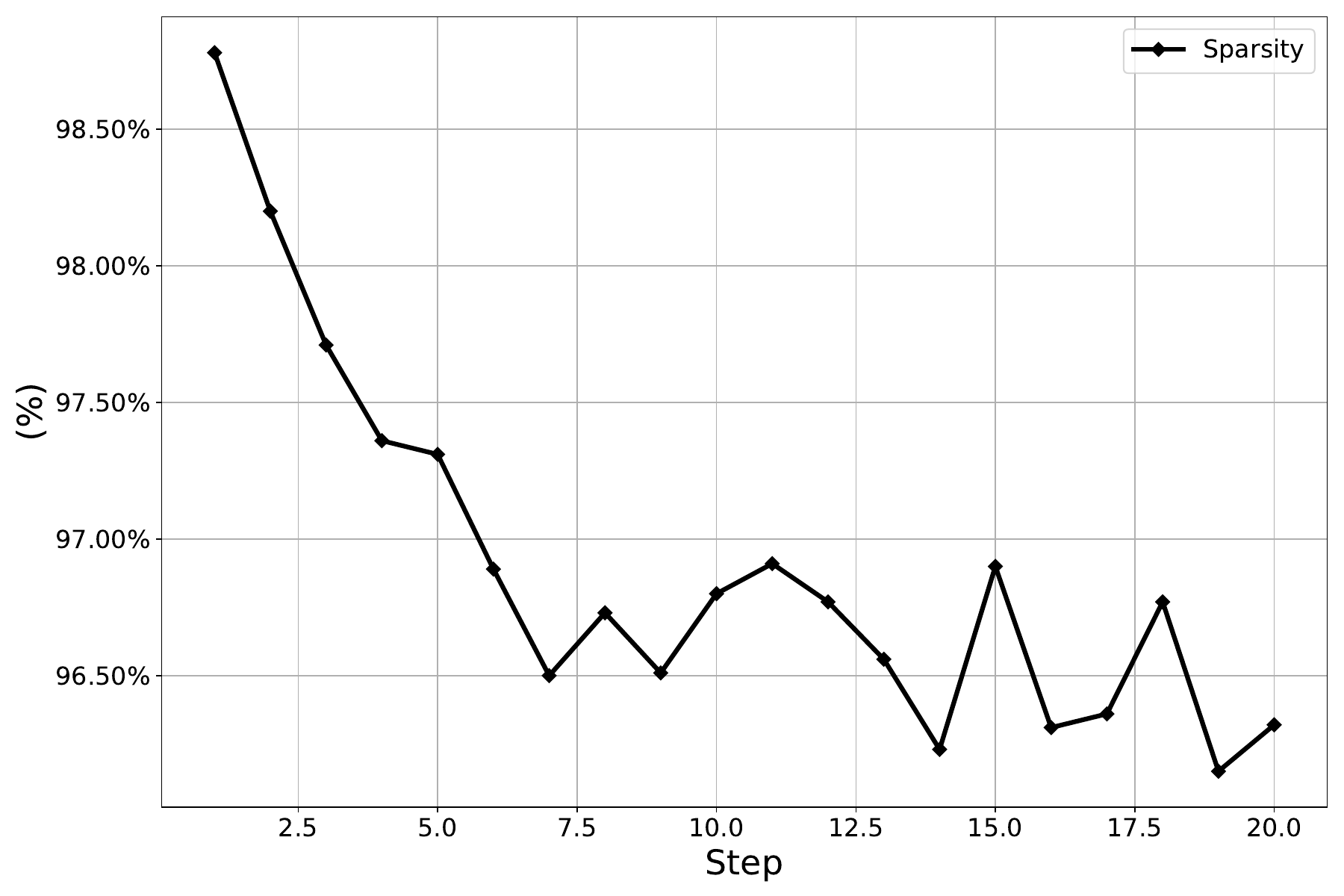}
        \caption{}
    \end{subfigure}
    \caption{Sparsity Analysis of intermediate checkpoints of \textsc{Prime} (a) shows the sparsity}
    \label{fig:intermediate_sparsity}
\end{figure}
Figure \ref{fig:intermediate_sparsity}(a) illustrates the sparsity of intermediate checkpoints alongside their accuracy on the MATH500 and AIME2024 tasks. The plot clearly demonstrates that all intermediate checkpoints exhibit non-trivial sparsity. Furthermore, as training progresses, the sparsity converges to a numerically significant asymptote, suggesting that a substantial proportion of weights remain unaffected even after prolonged period of training.
In Figure \ref{fig:intermediate_sparsity}(b), we report the sparsity of the $sparsity_{ij}$ for all consecutive checkpoint pairs, i.e., where $j = i+1$. Notably, in each successive step, on average, only 7\% of the weights receive a non-zero gradient update.

\section{Random Guessing baseline}
\label{sec:random_guessing}
If model 1 has sparsity $s_1$ and model 2 has sparsity $s_2$, the expected overlap is given by: \(\frac{(1 - s_1) \cdot (1 - s_2)}{100}\). Normalizing like earlier, we get \(\mathcal{O}_{1,random} = \frac{(1 - s_2)}{100}\) and \(\mathcal{O}_{2,random} = \frac{(1 - s_1)}{100}\). i.e. for random guessing the overlap for model 1 is the density of model 2. 

\end{document}